# Joint Label Prediction based Semi-Supervised Adaptive Concept Factorization for Robust Data Representation

Zhao Zhang, *Senior Member, IEEE,* Yan Zhang, Guangcan Liu, *Senior Member, IEEE,* Jinhui Tang, *Senior Member*, *IEEE,* Shuicheng Yan, *Fellow, IEEE*, and Meng Wang

**Abstract**— Constrained Concept Factorization (CCF) yields the enhanced representation ability over CF by incorporating label information as additional constraints, but it cannot classify and group unlabeled data appropriately. Minimizing the difference between the original data and its reconstruction directly can enable CCF to model a small noisy perturbation, but is not robust to gross sparse errors. Besides, CCF cannot preserve the manifold structures in new representation space explicitly, especially in an adaptive manner. In this paper, we propose a joint label prediction based Robust Semi-Supervised Adaptive Concept Factorization (RS²ACF) framework. To obtain robust representation, RS²ACF relaxes the factorization to make it simultaneously stable to small entrywise noise and robust to sparse errors. To enrich prior knowledge to enhance the discrimination, RS²ACF clearly uses class information of labeled data and more importantly propagates it to unlabeled data by jointly learning an explicit label indicator for unlabeled data. By the label indicator, RS²ACF can ensure the unlabeled data of the same predicted label to be mapped into the same class in feature space. Besides, RS²ACF incorporates the joint neighborhood reconstruction error over the new representations and predicted labels of both labeled and unlabeled data, so the manifold structures can be preserved explicitly and adaptively in the representation space and label space at the same time. Owing to the adaptive manner, the tricky process of determining the neighborhood size or kernel width can be avoided. Extensive results on public databases verify that our RS²ACF can deliver state-of-the-art data representation, compared with other related methods.

——————————— ◆ ———————————

## 1 INTRODUCTION

THE ever-increasing and enormous real data (e.g., visual images) of high-dimensional attributes have been posing challenges for the researchers working toward to handle the data representation issue. High-dimensional data representation is a fundamental problem in the areas of visual pattern recognition and data mining, because a "good" representation can discover the important latent structures and salient information in data for enhancing subsequent data clustering or classification. Matrix factorization is one of the representative representation learning methods of data, among which Principal Component Analysis (PCA) [1], Singular Value Decomposition (SVD) [2], Vector Quantization (VQ) [3], Non-negative Matrix Factorization (NMF) [4] and Concept Factorization (CF) [5] are several widely-used popular algorithms. Specifically, both NMF and CF enforce the resulting matrix factors to be nonnegative, which enables them to produce the parts-based representation of the original data [4-5][40-41].

————————————————
- Z. Zhang and Y. Zhang are with School of Computer Science and Technology, Soochow University, Suzhou 215006, China (e-mails: cszzhang@gmail.com, zhangyan0712suda@gmail.com)
- M. Wang and Z. Zhang are with School of Computer Science & School of Artificial Intelligence, Hefei University of Technology, Hefei, China. (e-mail: eric.mengwang@gmail.com)
- G. Liu is with Nanjing University of Information Science and Technology, Nanjing, China. (e-mail: gcliu@nuist.edu.cn)
- J. Tang is with School of Computer Science and Engineering, Nanjing University of Science and Technology, Nanjing 210094, China (e-mail: jinhuitang@njust.edu.cn)
- S. Yan is with Department of Electrical and Computer Engineering, National University of Singapore, Singapore (e-mail: eleyans@nus.edu.sg)

To the best of our knowledge, most existing factorization based methods aim at calculating the new basis vectors to represent data [4-5]. Specifically, NMF is to obtain two nonnegative factors $U$ and $V$ whose product can well approximate the data $X$, i.e., $X \approx UV^T$, where $U$ contains the basis vectors and $V^T$ is the new representation. The multiplicative updating rules minimizing NMF are given as

$$u_{ik}^{t+1} \leftarrow u_{ik}^t \frac{(XV)_{ik}}{(UV^TV)_{ik}}, \quad v_{jk}^{t+1} \leftarrow v_{jk}^t \frac{(X^TW)_{jk}}{(VU^TU)_{jk}}.$$

Note that existing studies on clustering and recognition have verified that NMF obtains the enhanced results than VQ, PCA and SVD [6]. More recently, several enhanced NMF variants have been proposed by extending NMF to the locality preserving scenario or discriminant scenario, e.g., Projective NMF (PNMF) [7], Graph Regularized NMF (GNMF) [8][27], Graph-based Discriminative NMF (GDNMF) with label information [44], Constrained NMF (CNMF) [9] and Semi-Supervised GNMF (SemiGNMF) [8]. Although the enhanced results have been delivered by aforementioned NMF variants, but NMF and its variants can only be performed in the original feature space of data points, so it cannot be executed in the reproducing kernel Hilbert space and the powerful kernel trick cannot be applied to NMF directly. Note that the reason will be presented after introducing the CF algorithm.

The recent CF model is a variation of NMF by expressing each cluster using a linear combination of data points and representing each data by a linear combination of the cluster centers [5]. As a result, CF can be performed either in the original space or in kernel space due to its flexible formulation, i.e., data $X$ is approximated by the product of three matrices $X$, $W$, and $V$, i.e., $X \approx XWV^T$. Representative variants of CF are Locally Consistent Concept Factorization (LCCF) [10] and Constrained Concept Factorization (CCF) [14]. The multiplicative updating rules minimizing CF are given in Eq.(3). One can find that the updating rules of CF clearly involve the inner product of $X$ and hence it can be easily kernelized. In contrast, the updating rules of NMF and its variants do not have the inner product of $X$, so they cannot be kernelized directly as CF. Note that kernelization is useful to extract nonlinear features hidden in the data by kernel-induced mapping. Besides, since the computation of those kernelized method mainly relies on the number of samples rather than the dimension, so the kernelized method will be applicable to handle the high-dimensional dataset.

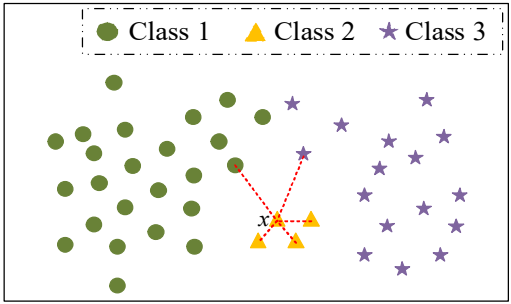

**Fig.1:** The nearest neighbor relationship of the example point $x$ when $k$ is set to 5, where the red dashed lines denote the nearest neighbor relationships.

It is worth noting that CF and its variants still have certain drawbacks. First, CF can only reveal the global Euclidean geometry of data, but fails to preserve the locality structures. To solve this issue, LCCF [10] uses the graph Laplacian to smooth the representation so that the local geometry information is encoded. Due to the local preservation ability, LCCF can obtain the enhanced and more descriptive manifold preserving representations than CF. Note that the manifold preservation ability is important for most representation learning tasks on the real datasets, since most human generated vision data (e.g., image and video) or non-vision data (e.g., text and document) are probably sampled from a sub-manifold of low intrinsic dimensionality, that is a topological space locally resembling the Euclidean space near each point, hidden in the high-dimensional Euclidean space [11][12][46][48]. For example, a set of face images of one individual changing smoothly from the front to side can form a sub-manifold. The closer the adjacent faces, the more similar, i.e., *locality*. Thus, it is important to preserve the manifold structures while learning the representations. But note that LCCF still suffers from two shortcomings. (1) LCCF determines the neighbors of each sample firstly by $k$-nearest neighbor search and fixes $k$ to construct the weights, but determining a suitable $k$ is rather tricky in reality. Also, using the same $k$ value artificially for each sample is also unreasonable, which does not consider the distribution of complex real data [11-13], such as the class-imbalance distribution, i.e., the scale of some classes is much larger than that of other classes. For example, in the fraudulent transactions identification, most transactions are normal, and only a small fraction of transactions are fraudulent. For such cases, LCCF suffers from an obvious drawback, especially for small scale classes, i.e., the neighbors of certain sample may be chosen from other classes due to the fixed number $k$. A simple class-imbalance example (3-class case) is given in Fig.1, and we use the sample $x$ of class 2 as an example. For this case, if we fix $k=5$ for LCCF to find the neighbors of $x$, two neighbors must be selected from other two classes (i.e., wrong connections), since class 2 only has four samples including $x$. The wrong connections can directly lead to inaccurate similarities and representation results. Thus, it is of great interest both in theory and in practice if we can extend CF to the adaptive weighting to determine $k$ automatically, i.e., adaptive for different real datasets. (2) LCCF pre-defines the weights and graph Laplacian independently prior to the factorization, but such an operation cannot ensure that the pre-encoded weights to be optimal for the subsequent representation. Thus, it would be better if we can incorporate the adaptive weighting into the factorization process further. Second, LCCF is also unsupervised as CF, i.e., they both cannot take advantage of class information of labeled data to improve the representation performance even if label information of samples is available. To solve this issue effectively, CCF [14] was recently proposed. CCF can obtain the representations of data consistent with known label information by defining an explicit label constraint matrix to represent the label information of labeled samples, and thus can ensure the original labeled data sharing the same label to be mapped into one class in low-dimensional space. But CCF cannot predict the labels of originally unlabeled data and also map them into their respective subspaces (i.e., clusters) in feature space, since it did not define an explicit label indicator matrix for unlabeled data, but simply sets it to be an identity matrix. Note that the constraint matrix contains the label indicator vectors for each sample, with the biggest entry in each label indicator vector determining the class assignment of the sample, as can be observed from Eq.(7). Because the number of labeled data is typically small in reality, the positive effects of incorporating the label constraints may be limited in CCF. Also, CCF cannot preserve the manifold structures in representation space explicitly, especially in an adaptive manner. But losing the adaptive local preservation power may also result in the degraded representations. Another common shortcoming of most CF based methods is that they directly minimize the difference between the data $X$ and its reconstruction $XWV^T$, which can only enable them to model the small noisy perturbation of the reconstruction, but they will fail in presence of gross sparse errors in reality.

In this paper, we therefore propose effective schemes to overcome the aforementioned shortcomings of LCCF and CCF, and at the same time inherit the advantages of them. The main contributions are shown as follows:

(1) Technically, A joint label prediction based partially labeled high-dimensional data representation framework, termed robust Semi-Supervised Adaptive Concept Factorization (RS²ACF), is proposed. To improve the representation ability, RS²ACF seamlessly incorporates the robust semi-supervised concept factorization, robust label prediction and the joint adaptive manifold preserving constraints on the label indicator and new representation into a unified framework. To obtain the robust representations, RS²ACF explicitly relaxes the factorization to make it simultaneously stable to small entrywise noise and robust to gross sparse errors. To interpret the resulting nonnegative factors from the factorization process of RS²ACF, we take the image recognition task as an example. Let $X$ be an image data, RS²ACF also approximates it by the product of several matrices $X$, $W$ and $V$, where $V^T=Z^TA^T$ is the new representation of $X$, $A$ is a label constraint matrix and $Z$ is an auxiliary matrix. Due to our unified model, the linear reconstruction $XW$ by the resulting factor $W$ can be interpreted as the cluster prototypes or basis vectors that discover the latent semantic structures hidden in the image data $X$. The other factor $V^T$ under the label constraint and adaptive manifold preserving constraint can be interpreted as the discriminative adaptive locality preserving new representations or coordinates of the original data in $X$, which exhibits attractive properties over CCF and LCCF. Besides, the dimension of learnt new representation $V^T$ is usually much smaller than that of $X$ (i.e., compact representation of the original high-dimensional data) in practical applications, which therefore can facilitate other subsequent data mining tasks, e.g., clustering and classification of high-dimensional real-world data.

(2) To enable RS²ACF to deal with the complex or special distributions potentially, such as the class-imbalance distribution, we present an adaptive strategy to preserve the local manifold structures in the factorization process. That is, RS²ACF integrates the adaptive weighting and the semi-supervised concept factorization seamlessly into a unified model. Specifically, RS²ACF includes a neighborhood reconstruction error encoded by sharing the graph weight matrix in original space, new representation space and label space at the same time for joint minimization, so the manifold structures of labeled and unlabeled data can be preserved explicitly and adaptively in the learnt representation space and label space. That is, we do not need to specify the number $k$ of neighbors at all, since the neighbors of each data point are determined automatically via minimizing the reconstruction error jointly. By updating the weights, new representation and predicted labels alternately, one can ensure the weights to be optimal for the data representation and can also reduce the wrong inter-class connections clearly as can be seen from the examples in Figs.2 and 3, while existing locality based LCCF cannot ensure such issue. By the adaptive weighting, the tricky issue of determining the optimal neighborhood size $k$ suffered in LCCF can be avoided, i.e., RS²ACF can be adaptive to different distributions of real datasets, which can make it more applicable to the real applications.

(3) To effectively improve the discriminating power of learnt representation, our RS²ACF clearly considers making full use of the class information of labeled data and enriching the supervised prior information. Specifically, RS²ACF considers propagating the class information of labeled data to unlabeled data and further predict the labels of those unlabeled data by jointly learning a robust label predictor $P$ and an explicit label indicator matrix for the unlabeled data. By the predicted labels of unlabeled samples, RS²ACF can explicitly ensure the unlabeled data sharing the same predicted label to be mapped into their respective subspaces in feature space. Thus, the discriminating ability of the new representations can be potentially enhanced compared with CCF, since CCF only defines a label indicator sub-matrix for labeled data and can only ensure those originally labeled data of the same class to be mapped into respective subspaces. As a result, RS²ACF will be more applicable to deal with the semi-supervised learning task that the number of labeled samples is limited, which is often encountered in real applications. The sparse L2,1-norm is also used on the predictor $P$ so that the embedded discriminative soft labels can be obtained in the projective latent subspace for classification [15-17].

The paper is outlined as follows. Section 2 reviews the related work briefly. Section 3 presents the formulation and optimization of RS²ACF mathematically. In Section 4, we show the connections between RS²ACF and other related work. Section 5 shows simulation results on public datasets. Finally, the paper is concluded in Section 6.

## 2 RELATED WORK

We briefly review CF [5], LCCF [10] and CCF [14] here.

### 2.1 Concept Factorization (CF)

Concept Factorization is a classical unsupervised matrix factorization method for data representation. Given a data matrix $X=[x_1,x_2,...,x_N]\in \mathbb{R}^{D\times N}$, where $x_i, i\in 1,2,...,N$ is a sample vector, $N$ is the number of samples and $D$ is the original dimension of each sample. Letting $U\in\mathbb{R}^{D\times r}$ and $V^T\in\mathbb{R}^{r\times N}$ be two nonnegative matrices whose product $UV^T\in\mathbb{R}^{D\times N}$ is the approximation to the original data $X$, where the rank $r$ is a constant, by representing each basis by using a nonnegative linear combination of $x_i$, that is, $\sum_{i=1}^N w_{ij}x_i$, where $w_{ij}\geq 0$, then CF aims at calculating the approximate relation as $X\approx XWV^T$. That is, CF proposes to solve the following minimization problem:

$$O=\left\|X-XWV^T\right\|^2, \tag{1}$$

where $W=[w_{ij}]\in\mathbb{R}^{N\times r}$, $XW$ approximates the bases, $V^T$ is the new representation, and $V^T$ is the transpose of $V$. The multiplicative updating rules of CF are described as

$$w_{ik}^{t+1}\leftarrow w_{ik}^t\frac{(KV)_{ik}}{(KWV^TV)_{ik}}, \quad v_{jk}^{t+1}\leftarrow v_{jk}^t\frac{(KW)_{jk}}{(VW^TKW)_{jk}}, \tag{2}$$

where $K=X^TX$ is the inner product matrix or kernel matrix. After the convergence of the above rules, the new representation $V^T$ of the original data $X$ can be obtained.

### 2.2 Locally Consistent Concept Factorization (LCCF)

LCCF learns the manifold preserving new representations of original data by including a geometrically based regularizer. Specifically, LCCF firstly constructs a graph $G(R, E)$ with $N$ nodes over the samples in $X$, where each vertex $r_i$ in vertex set $R$ corresponds to a sample $x_i$, and the edge weight $S_{ij}$ connecting $x_i$ and $x_j$ can be defined as

$$S_{ij}=\begin{cases} x_i^T x_j/(\|x_i\|\|x_j\|), & \text{if } x_i\in N_k(x_j) \text{ or } x_j\in N_k(x_i) \\ 0, & \text{otherwise} \end{cases}, \tag{3}$$

where $N_k(x_i)$ is the set including the $k$ nearest neighbors of $x_i$. The regularization term $\Re$ can then be defined as

$$\Re = 1/2 \sum_{i,j=1}^{N} \|v_i - v_j\|^2 S_{ij} = tr(V^T L V), \quad (4)$$

where $v_i$ is the $i$-th column of $V$, graph Laplacian $L=D-S$, $D$ is a diagonal matrix whose entries are column (or row, since $S$ is symmetric) sums of $S$, i.e., $D_{ii} = \sum_j S_{ij}$. Thus, LCCF solves the following objective function:

$$\min_{W,V} \|X - XWV^T\|^2 + \lambda tr(V^T L V), \quad s.t.\ W, V \geq 0, \quad (5)$$

where $\lambda \geq 0$ is a regularization factor. Finally, the updating rules of $W$ and $V$ can be computed as [10]:

$$w_{ik}^{t+1} \leftarrow w_{ik}^t \frac{(KV)_{ik}}{(KWV^T V)_{ik}}, \quad v_{jk}^{t+1} \leftarrow v_{jk}^t \frac{(KW + \lambda SV)_{jk}}{(VW^T KW + \lambda DV)_{jk}}. \quad (6)$$

### 2.3 Constrained Concept Factorization (CCF)

CCF extends CF to semi-supervised scenario [33][45][47-48] by using the class information of labeled data as additional constraint. Supposing that the dataset $X$ contains a labeled set $X_L \in \mathbb{R}^{D \times l}$ and an unlabeled set $X_U \in \mathbb{R}^{D \times u}$, that is, $l+u=N$ and $X=[X_L, X_U] \in \mathbb{R}^{D \times (l+u)}$, where $l$ and $u$ are the numbers of labeled and unlabeled data respectively, then CCF can drive the constrained factorization by representing label information by a label constraint matrix $A$. Let $A_L \in \mathbb{R}^{l \times c}$ be a class indicator matrix defined over class information of $X_L$, where $c$ is the number of classes. The entry $(A_L)_{ij}$ is defined as 1 if $x_i$ is labeled with the $j$-th class, and otherwise $(A_L)_{ij} = 0$. Note that CCF does not define an explicit label indicator matrix for $X_U$ and simply uses an $u \times u$ identity matrix $I_{u \times u}$ for $X_U$. Thus, the overall label constraint matrix $A$ is defined as

$$A = \begin{bmatrix} (A_L)_{l \times c} & 0 \\ 0 & I_{u \times u} \end{bmatrix} \in \mathbb{R}^{(l+u) \times (c+u)}. \quad (7)$$

To ensure the points sharing the same label are mapped into the same class in low-dimensional space (i.e., same $v_i$), CCF imposes the label constraint by an auxiliary matrix $Z$:

$$V = AZ. \quad (8)$$

By substituting $V=AZ$ into CF, CCF finds a non-negative matrix $W \in \mathbb{R}^{N \times r}$ and $Z \in \mathbb{R}^{(c+u) \times r}$ from

$$O = \|X - XWZ^T A^T\|^2. \quad (9)$$

The updating rules for $W$ and $Z$ are described as [14]:

$$w_{ik}^{t+1} \leftarrow w_{ik}^t \frac{(KAZ)_{ik}}{(KWZ^T A^T AZ)_{ik}}, \quad z_{jk}^{t+1} \leftarrow z_{jk}^t \frac{(A^T KW)_{jk}}{(A^T AZW^T KW)_{jk}}. \quad (10)$$

## 3 ROBUST SEMI-SUPERVISED ADAPTIVE CONCEPT FACTORIZATION (RS²ACF)

### 3.1 The Objective Function

We describe the idea and formulation of our RS²ACF that overcomes the shortcomings of both CCF and LCCF to improve the representation and discriminating abilities. Given a partially labeled dataset $X=[X_L, X_U] \in \mathbb{R}^{D \times (l+u)}$, our RS²ACF incorporates the semi-supervised concept factorization, robust label prediction and joint adaptive manifold preserving constraints on the label indicator and new representation seamlessly into a unified model. RS²ACF also considers enhancing the robust properties of factorization process. Specifically, our RS²ACF relaxes the original reconstruction error $\|E_{ori}\|_F^2$, where $E_{ori} = X - XWZ^T A^T$ in CCF, to $E_{ori} = X - E - XWZ^T A^T$, where $E$ is the L2,1-norm regularized sparse matrix. Thus, RS²ACF can be simultaneously stable to small entrywise noise that is modeled by $\|E_{ori}\|_F^2$ and robust to the gross sparse errors modeled by L2,1-norm [15-16] based $E$, i.e., $\|E^T\|_{2,1}$. Thus, the recovered low-dimensional new representation of original data can be potentially more accurate. Finally, our RS²ACF can jointly calculate a low-dimensional new representation of original data $X$, an auxiliary matrix $Z \in \mathbb{R}^{2c \times r}$, an adaptive weighting matrix $Q \in \mathbb{R}^{(l+u) \times (l+u)}$, a sparse term $E$ encoding the sparse errors, a label indicator $A_U \in \mathbb{R}^{u \times c}$ for unlabeled samples in $X_U$ and a robust label predictor $P \in \mathbb{R}^{D \times c}$ for the class assignment. These can lead to the following initial problem for our presented RS²ACF:

$$\min_{\substack{W,Z,E, \\ Q,A_U,P}} \|X - E - XWZ^T A^T\|_F^2 + \gamma \|E^T\|_{2,1} + \alpha f(Q) + \beta g(A_U, P) \\ s.t.\ W, Z, Q \geq 0,\ Q_{ii} = 0 \quad , (11)$$

where $\|X - E - XWZ^T A^T\|_F^2$ is the robust reconstruction error over all data, $f(Q)$ is the adaptive locality constraint term of representations, and $g(A_U, P)$ is the joint learning term of the class indicator for unlabeled data and robust label predictor. $W, Z, Q \geq 0$ are the non-negative constraints, and $Q_{ii} = 0$ is to avoid the trivial solution $Q=I$, where $I$ is the identity matrix. $\alpha \geq 0$ and $\beta \geq 0$ are two parameters. Our RS²ACF defines the overall fully labeled constraint matrix $A$, auxiliary matrix $Z$ and nonnegative matrix $W$ as

$$A = \begin{bmatrix} A_L & 0 \\ 0 & A_U \end{bmatrix} \in \mathbb{R}^{(l+u) \times (c+c)},\ A_L \in \mathbb{R}^{l \times c},\ A_U \in \mathbb{R}^{u \times c}, \quad (12)$$

$$Z = \begin{bmatrix} Z_L \\ Z_U \end{bmatrix} \in \mathbb{R}^{(c+c) \times r},\ W = \begin{bmatrix} W_L \\ W_U \end{bmatrix} \in \mathbb{R}^{(l+u) \times r}, \quad (13)$$

where $A_L$ is the class indicator for labeled data. Note that RS²ACF also learns an explicit class indicator $A_U$ for unlabeled data $X_U$ so that the new representations of both labeled and unlabeled data can be well grouped in feature space. Although the term $\|X - E - XWZ^T A^T\|_F^2$ in RS²ACF shares the similar form as that of CCF, RS²ACF can enable the model to be robust to small entrywise noise and gross sparse errors jointly and can estimate the labels of unlabeled data by $P$. The definitions of $Z$ and $A$ are also different for CCF and RS²ACF. Next, we describe $f(Q)$ and $g(A_U, P)$ for the adaptive weighting and classification.

#### 1) Adaptive Locality Constraint of Representations

To keep the local manifold structures of new representation, RS²ACF obtains the adaptive reconstruction weight matrix $Q$ from the following function:

$$f(Q) = \|X - XQ\|_F^2 + \|Z^T A^T - Z^T A^T Q\|_F^2 + \|P^T X - P^T XQ\|_F^2, \quad (14) \\ s.t.\ Q \geq 0,\ Q_{ii} = 0$$

where $AZ$ is the low-dimensional representation of $X$, $P$

is a linear projection classifier that will be discussed shortly, i.e., $P^T X$ are the predicted labels of $X$. Clearly, the weight matrix $Q$ minimizes the reconstruction errors over original data $X$, new representations $AZ$ and estimated labels $P^T X$ jointly. That is, the neighborhood encoded by weights $Q$ are consistently shared in the original space, new representation space and label space at the same time. Thus, the local manifold structures can be clearly preserved in representation space and label space.

### 2) Joint Learning of Class Indicator $A_u$ for $X_u$ and Robust Label Predictor $P$

We discuss how to obtain an explicit class indicator $A_u$ for unlabeled data by jointly computing a robust label predictor $P$. Different from CCF whose positive effects of incorporating the label constraints may be limited, we would like to investigate how to estimate the class information of originally unlabeled data so that all the training data (including both labeled and unlabeled data) can be grouped and represented more accurately. To this end, we propose to define the following embedding based function to compute the class indicator $A_u$ for unlabeled data:

$$g(A_U, P) = \|A_L - X_L^T P\|_F^2 + \|A_U - X_U^T P\|_F^2 + \|P\|_{2,1}, \quad (15)$$

from which it is clear that the embedded labels $X_L^T P$ and $X_U^T P$ only depends on the indicators $A_L$ and $A_U$. $\|P\|_{2,1}$ is L2,1-norm based label predictor that can propagate class information of labeled data to unlabeled data. Note that the sparse L2,1-norm can ensure the robust properties of the predictor $P$ so that it is robust to noise and outliers in data [15-16], and it can also enable the discriminating soft labels to be predicted in the latent label subspace. To clarify the advantages of using $f(Q)$ and $g(A_U, P)$ more clearly, we discuss the sum of them, which is given as

$$\alpha f(Q) + \beta g(A_U, P)$$
$$= \alpha \left( \|X - XQ\|_F^2 + \|Z^T A^T - Z^T A^T Q\|_F^2 + \|P^T X - P^T XQ\|_F^2 \right),$$
$$+ \beta \left( \|A_L - X_L^T P\|_F^2 + \|A_U - X_U^T P\|_F^2 + \|P\|_{2,1} \right)$$

where the minimization of $\beta \left( \|A_L - X_L^T P\|_F^2 + \|A_U - X_U^T P\|_F^2 \right) + \alpha \|P^T X - P^T XQ\|_F^2$ can clearly mean that the process of predicting the labels $X_U^T P$ of the unlabeled samples not only replies on the class indicator $A_U$, but also receives partial information from the adaptive neighborhood of each data in terms of label reconstruction by $\|P^T X - P^T XQ\|_F^2$ [17][36], where the adaptive neighborhood is mainly encoded by the adaptive weights in $Q$. As a result, the predicted labels of the unlabeled data will be potentially more accurate for classification, especially for the cases that the estimated class indicator $A_U$ is not very reliable in special cases.

Thus, the final objective function of RS²ACF is given as

$$\min_{W,Z,E,Q,A_U,P} \|X - E - XWZ^T A^T\|_F^2 + \gamma \|E\|_{2,1}$$
$$+ \alpha \left( \|X - XQ\|_F^2 + \|Z^T A^T - Z^T A^T Q\|_F^2 + \|P^T X - P^T XQ\|_F^2 \right). \quad (16)$$
$$+ \beta \left( \|A_L - X_L^T P\|_F^2 + \|A_U - X_U^T P\|_F^2 + \|P\|_{2,1} \right)$$
$$s.t.\ W,Z,Q \geq 0,\ Q_{ii} = 0$$

Note that we can obtain a weight matrix $Q$, a class indicator $A_u$ for unlabeled data, and a linear projection classifier $P$ jointly from Eq.(16). Thus, $XWZ^T A^T$ in our RS²ACF can be regarded as the robust semi-supervised adaptive neighborhood preserving representation of original data. In what follows, we detail the optimization procedures.

### 3.2 Optimization

We show how to solve the objective function of RS²ACF in Eq.(16). Since the variables, i.e., $W$, $Z$, $A_u$, $Q$ and $P$, rely on each other, they cannot be solved directly. Thus, we present an alternate method to obtain the local optima of the model by iterative updating rules [8], i.e., we update one variable each time by fixing others. The optimization can be alternately performed by the following steps:

**1) Fix others, update the matrix factor $W$:**

We can update the variable $W$ with other variables given. By removing terms that are independent on $W$, we can have the following reduced formulation:

$$\min_W \|X - E - XWZ^T A^T\|_F^2,\ s.t.\ W \geq 0. \quad (17)$$

Define $K_E = (X - E)^T (X - E)$ and by using the property of matrix, the above problem can be reformulated as

$$\min_W J(W) = tr\left( (X - E - XWZ^T A^T)^T (X - E - XWZ^T A^T) \right)$$
$$= tr\left( K_E - 2(X - E)^T XWZ^T A^T + AZW^T KWZ^T A^T \right), \quad (18)$$

Let $\psi_{ik}$ be the Lagrange multiplier for the constraint $w_{ik} \geq 0$ and $\Psi = [\psi_{ik}]$, then the Lagrange function $\mathcal{L}_1$ of the above problem can be constructed as

$$\mathcal{L}_1 = tr\left( K_E - 2(X - E)^T XWZ^T A^T + AZW^T KWZ^T A^T \right) + tr(\Psi W^T). \quad (19)$$

The partial derivative of $\mathcal{L}_1$ w.r.t. $W$ can be computed as

$$\partial \mathcal{L}_1 / \partial W = 2KWZ^T A^T AZ - 2X^T (X - E) AZ + \Psi. \quad (20)$$

By applying the Karush-Kuhn-Tucker (KKT) condition $\psi_{ik} w_{ik} = 0$, we can obtain the following equation:

$$(KWZ^T A^T AZ)_{ik} w_{ik} - (X^T (X - E) AZ)_{ik} w_{ik} = 0. \quad (21)$$

Thus, we obtain the updating rule of $W = [w_{ik}]$ as

$$w_{ik} \leftarrow w_{ik} \frac{(X^T (X - E) AZ)_{ik}}{(KWZ^T A^T AZ)_{ik}}. \quad (22)$$

**2) Fix others, update matrix $Z$ and class indicator $A_u$:**

We update variables $Z$ and $A_u$ in this step. Note that we update the sub-matrices $Z_L$ and $Z_U$ separately. Based on the matrix property, that is, $\|A\|_F^2 = tr(A^T A) = tr(AA^T)$, by removing terms that are irrelevant to $A_u$, $Z_L$ and $Z_U$, and expanding $\|X - E - XWZ^T A^T\|_F^2$ into labeled and unlabeled parts, we can have the following reduced formulation:

$$\min_{Z_L, Z_U, A_U} J(Z_L, Z_U, A_U) = \|X_L - E_L - X_L W_L Z_L^T A_L^T\|_F^2$$
$$+ \|X_U - E_U - X_U W_U Z_U^T A_U^T\|_F^2 + \alpha tr(Z^T A^T HAZ), \quad (23)$$
$$+ \beta \|A_U - X_U^T P\|_F^2,\ s.t.\ Z_L, Z_U \geq 0$$

where $H = (I - Q)(I - Q)^T$ is an auxiliary matrix based on the adaptive weight matrix $Q$, and can be further splitted into four blocks over the labeled and unlabeled data:

$$H = \begin{bmatrix} H_{LL} & H_{LU} \\ H_{UL} & H_{UU} \end{bmatrix}, \quad (24)$$

To simplify the optimization, we first expand the semi-supervised concept factorization term $tr(Z^T A^T HAZ)$ as

$$tr(Z^T A^T HAZ)$$
$$= tr\left( [Z_L^T, Z_U^T] \begin{bmatrix} A_L^T & 0 \\ 0 & A_U^T \end{bmatrix} \begin{bmatrix} H_{LL} & H_{LU} \\ H_{UL} & H_{UU} \end{bmatrix} \begin{bmatrix} A_L & 0 \\ 0 & A_U \end{bmatrix} \begin{bmatrix} Z_L \\ Z_U \end{bmatrix} \right)$$
$$= tr\left( [Z_L^T A_L^T, Z_U^T A_U^T] \begin{bmatrix} H_{LL} & H_{LU} \\ H_{UL} & H_{UU} \end{bmatrix} \begin{bmatrix} A_L Z_L \\ A_U Z_U \end{bmatrix} \right) \quad .(25)$$
$$= tr(Z_L^T A_L^T H_{LL} A_L Z_L + Z_U^T A_U^T H_{UL} A_L Z_L)$$
$$+ tr(Z_L^T A_L^T H_{LU} A_U Z_U + Z_U^T A_U^T H_{UU} A_U Z_U)$$

Let $\phi_{ik}$, $\tau_{ik}$, and $\varepsilon_{ik}$ be the Lagrange multipliers for the constraints $(z_L)_{ik} \geq 0$, $(z_U)_{ik} \geq 0$, $(a_U)_{ik} \geq 0$, and $\Phi = [\phi_{ik}]$, $T = [\tau_{ik}]$, $E = [\varepsilon_{ik}]$ respectively, the Lagrange function $\mathcal{L}$ of the problem in Eq.(23) can then be constructed as

$$\mathcal{L} = J(Z_L, Z_U, A_U) + tr(\Phi Z_L^T) + tr(T Z_U^T) + tr(E A_U^T). \quad (26)$$

Let $K_U = (X_U - E_U)^T X_U$ and $K_L = (X_L - E_L)^T X_L$, the partial derivatives w.r.t. $Z_L, Z_U$ and $A_U$ can be defined as

$$\frac{\partial \mathcal{L}}{\partial Z_L} = -2 A_L K_L W_L + 2 M_L + 2\alpha O_L + \Phi, \quad (27)$$

$$\frac{\partial \mathcal{L}}{\partial Z_U} = -2 A_U K_U W_U + 2 M_U + 2\alpha O_U + T, \quad (28)$$

$$\frac{\partial \mathcal{L}}{\partial A_U} = -2 K_U W_U Z_U^T + 2 G_U + 2\alpha G_L + 2\beta(A_U - X_U^T P) + E, \quad (29)$$

where $M_L = A_L^T A_L Z_L W_L^T X_L^T X_L W_L$, $O_L = A_L^T H_{LL} A_L Z_L + A_L^T H_{LU} A_U Z_U$, $M_U = A_U^T A_U Z_U W_U^T X_U^T X_U W_U$, $O_U = A_U^T H_{UU} A_U Z_U + A_U^T H_{LU}^T A_L Z_L$, $G_U = A_U Z_U W_U^T X_U^T X_U W_U Z_U^T$, $G_L = H_{UU} A_U Z_U Z_U^T + H_{LU}^T A_L Z_L Z_U^T$. By applying the KKT conditions $\phi_{ik}(z_U)_{ik} = 0$, $\tau_{ik}(z_L)_{ik} = 0$, and $\varepsilon_{ik}(a_U)_{ik} = 0$, we can obtain the following equations:

$$(-A_L K_L W_L)_{ik}(z_L)_{ik} + (M_L)_{ik}(z_L)_{ik} + \alpha(O_L)_{ik}(z_L)_{ik} = 0, \quad (30)$$

$$(-A_U K_U W_U)_{ik}(z_U)_{ik} + (M_U)_{ik}(z_U)_{ik} + \alpha(O_U)_{ik}(z_U)_{ik} = 0, \quad (31)$$

$$(-K_U W_U Z_U^T)_{ik}(a_U)_{ik} + (G_U)_{ik}(a_U)_{ik} + \alpha(G_L)_{ik}(a_U)_{ik} + \beta(A_U - X_U^T P)_{ik}(a_U)_{ik} = 0 \quad .(32)$$

Thus, the following updating rules for $Z_L$, $Z_U$, and $A_U$ can be easily obtained from Eqs.(30-32):

$$(z_L)_{ik} \leftarrow (z_L)_{ik} \frac{(A_L K_L W_L)_{ik}}{(M_L + \alpha O_L)_{ik}}, \quad (33)$$

$$(z_U)_{ik} \leftarrow (z_U)_{ik} \frac{(A_U K_U W_U)_{ik}}{(M_U + \alpha O_U)_{ik}}, \quad (34)$$

$$(a_U)_{ik} \leftarrow (a_U)_{ik} \frac{(\beta X_U^T P + K_U W_U Z_U^T)_{ik}}{(G_U + \alpha G_L + \beta A_U)_{ik}}. \quad (35)$$

**3) Fix others, update the sparse error term $E$:**

To update $E$, we have the following reduced problem by removing the irrelevant terms from the objective function:

$$\min_E J(E) = \|X - E - XWZ^T A^T\|_F^2 + \gamma \|E^T\|_{2,1}. \quad (36)$$

By the properties of L2,1-norm [15-17], $\|E\|_{2,1} = 2tr(E^T \Im E)$, where $\Im \in \mathbb{R}^{N \times N}$ is a diagonal matrix with the entries being $\omega_{ii} = 1/\left[2\|E_i^T\|_2\right]$, $i = 1, \ldots, N$, where $E_i$ is the $i$-th column of $E$. When each $E_i \neq 0$, we have an approximate problem:

$$\min_E J(E) = tr((\aleph - E)(\aleph - E)^T) + \gamma tr(E \Im E^T). \quad (37)$$

where $\aleph = X - XWZ^T A^T$. By taking the derivative of $J(E)$ w.r.t. $E$, one can obtain the updating rule for $E$ as

$$E = (X - XWZ^T A^T)(I + \gamma \Im)^{-1}. \quad (38)$$

Then, we can update $\Im$ as $\Im = diag(\omega_{ii})$, $\omega_{ii} = 1/\left[2\|E_i^T\|_2\right]$.

**4) Fix others, update adaptive weighting matrix $Q$:**

We fix $Z$ and $A_U$ to update the weight matrix $Q$. Note that we regard $Z$ and $A$ as a whole in the step of updating $Q$. We can have the following reduced problem by removing the irrelevant terms from the objective function:

$$\min_Q J(Q) = \alpha\left(\|X - XQ\|_F^2 + \|Z^T A^T - Z^T A^T Q\|_F^2 + \|P^T X - P^T XQ\|_F^2\right)$$
$$= \left\| \begin{pmatrix} \sqrt{\alpha} X \\ \sqrt{\alpha} Z^T A^T \\ \sqrt{\alpha} PX \end{pmatrix} - \begin{pmatrix} \sqrt{\alpha} X \\ \sqrt{\alpha} Z^T A^T \\ \sqrt{\alpha} PX \end{pmatrix} Q \right\|_F^2, \text{ s.t. } Q \geq 0, Q_{ii} = 0$$
$$(39)$$

Let $Y_{new} = \left(\sqrt{\alpha} X^T, \sqrt{\alpha} AZ, \sqrt{\alpha} X^T P^T\right)^T$, then the optimization of Eq.(36) is equivalent to solving the following one:

$$\min_Q J(Q) = \left(\|Y_{new} - Y_{new} Q\|_F^2\right), \text{ s.t. } Q \geq 0, Q_{ii} = 0. \quad (40)$$

Similarly, we can define the Lagrange function $\mathcal{L}_2$ of $Q$ as

$$\mathcal{L}_2 = tr\left(Y_{new}(I - Q)(I - Q)^T Y_{new}^T\right) + tr(\Gamma Q^T), \quad (41)$$

where $\Gamma = [\gamma_{ik}]$ is Lagrange multiplier for constraint $q_{ik} \geq 0$. Then, the derivative of $\mathcal{L}_2$ w.r.t. $Q$ can be obtained as

$$\partial \mathcal{L}_2 / \partial Q = 2 Y_{new}^T Y_{new} Q - 2 Y_{new}^T Y_{new} + \Gamma. \quad (42)$$

Since $\gamma_{ik} q_{ik} = 0$, according to the KKT condition, we have

$$\left(Y_{new}^T Y_{new} Q\right)_{ik} q_{ik} - \left(Y_{new}^T Y_{new}\right)_{ik} q_{ik} = 0. \quad (43)$$

Finally, we obtain the updating rule for weights $Q$ as

$$q_{ik} \leftarrow q_{ik} \frac{\left(Y_{new}^T Y_{new}\right)_{ik}}{\left(Y_{new}^T Y_{new} Q\right)_{ik}}. \quad (44)$$

After the reconstruction weight matrix $Q$ is updated by the above equation, we further make $Q_{ii} = 0$.

**4) Given $A_U$, update the robust label predictor $P$:**

The predicted label matrix $A_U$ of unlabeled data is fixed in this step, and we show how to update the label predictor $P$. In this case, we have the following reduced problem:

$$\min_P J(P) = \alpha \|P^T X - P^T XQ\|_F^2$$
$$+ \beta\left(\|A_L - X_L^T P\|_F^2 + \|A_U - X_U^T P\|_F^2 + \|P\|_{2,1}\right). \quad (45)$$

Let $B \in \mathbb{R}^{D \times D}$ be a diagonal matrix with the diagonal en-tries being $b_{ii} = 1/\left[2\|p^i\|_2\right]$, $i=1,...,D$, then we can transform Eq.(45) into the following matrix trace form:

$$\min_P J(P) = \alpha tr(P^T XHX^T P) + \beta tr\left((A_L - X_L^T P)^T (A_L - X_L^T P)\right)$$
$$+ \beta tr\left((A_U - X_U^T P)^T (A_U - X_U^T P)\right) + \beta tr(P^T BP) ,$$
(46)

when each $p^i \neq 0$, $i=1,...,D$, where $p^i$ is the $i$-th row vector of the label predictor $P$. By taking the derivative of $J(P)$ w.r.t. $P$, one can obtain the updating rule for $P$ as

$$P = \beta\left(\alpha XHX^T + \beta X_L X_L^T + \beta X_U X_U^T + \beta B\right)^{-1}(X_L A_L + X_U A_U) .$$
(47)

In the $t$-th iteration, after $P_t$ is updated, we can easily update the diagonal matrix $B_t$ as

$$B_t = diag(b_{ii}^t), b_{ii}^t = 1/\left[2\|p_t^i\|_2\right], i=1,2,...,D ,$$
(48)

where $p_t^i$ is the $i$-th row vector of the label predictor $P_t$. To present our method completely and clearly, we summarize the optimization procedures of our RS²ACF in Algorithm 1, where the diagonal matrix $\Im$ and $B$ is initialized as identity matrices as [16-17] so that $E_i^T \neq 0$ and $p^i \neq 0$ are satisfied during the iterative process. Note that minimizing $\|A_L - X_L^T P\|_F^2 + \|A_U - X_U^T P\|_F^2$ means that the prediction results by the classifier $P$ depend on the class indicators $A_L$ and $A_U$. Since $A_L$ is defined directly based on class information of labeled data, it is usually accurate. In contrast, the class indicator $A_U$ is defined for unlabeled data, but unlabeled data has no supervised prior information. Thus, $A_U$ is often initialized as a matrix of all zeros or an identity matrix in CCF [14], but minimizing $\|A_U - X_U^T P\|_F^2$ jointly with these initializations tend to result in inaccurate or even wrong predicted labels $X_U^T P$ over unlabeled data, i.e., the projection $P$ cannot ensure to be a label predictor. In other words, the initialization of $A_U$ is very important in our formulation. To address the above issues, two effective strategies are used to ensure that the projection $P$ to be a label predictor and the prediction labels $X_U^T P$ to be accurate as much as possible in this present paper. First, since $A_L$ is accurate, we can initialize the label predictor $P$ by $\min_P \|A_L - X_L^T P\|_F^2 + \|P\|_F^2$ with the solution $P = (X_L X_L^T + I)^{-1} X_L A_L$. Based on the initialized $P$, we can initialize the class indicator $A_U$ for unlabeled data further as $X_U^T P$. Second, the manifold preserving regularization $\|P^T X - P^T XQ\|_F^2$ can preserve the relationships between the labeled and unlabeled data so that the process of predicting the labels of unlabeled data can receive partial information from $A_U$ and also partly come from the neighborhoods [36] so that the predictions can be more accurate.

### 3.3 Convergence Analysis

We present the convergence analysis of our RS²ACF in this section. Specifically, we have the following theorem (i.e., Theorem 1) regarding the above iterative updating rules. Theorem 1 can ensure the convergence of the iterations and thus the final solution will be a local optimum.

**Theorem 1:** The objective function of our RS²ACF method in Eq.(16) is non-increasing in the presented updating rules of Eqs.(22) (33-35) and (44).

---

**Algorithm 1: Our Proposed RS²ACF Framework**
**Inputs:** Data matrix $X = [X_L, X_U] \in \mathbb{R}^{D \times (l+u)}$, positive constant $r$, and hyperparameters $\alpha, \beta$;
**Initialization:** $t = 0$; Initialize $\Im$ and $B$ to be identity matrices; Initialize $W$ and $Z$ as random matrices; Initialize the error $E$ to be zero matrix; Initialize the label predictor $P$ as $P = (X_L X_L^T + I)^{-1} X_L A_L$ and the class indicator $A_U$ for unlabeled data as $X_U^T P$. Initialize the entries of the weight matrix $Q$ by the cosine similarity, i.e., $Q_{ij} = \cos(x_i, x_j)$;
**While not converged do**
1. Update $Z_L$, $Z_U$ and $A_U$ by Eqs.(33-35). Obtain the whole constraint matrix $A$ by Eq.(12) and $Z=[Z_L; Z_U]$;
2. Update the linear label predictor $P$ by Eq. (44);
3. Update the factorization matrix $W$ by Eq.(22);
4. Update the sparse error matrix $E$ by Eq.(38);
5. Update the adaptive weighting matrix $Q$ by Eq. (44);
6. Update the diagonal matrices $\Im$ and $B$ accordingly;
7. Convergence check: if $\|\mathcal{O}^t - \mathcal{O}^{t+1}\| \leq 10^{-4}$, stop; else, return to step 1, where $\mathcal{O}$ is the objective function value.
**Output:** New robust representation $A^*Z^*$ of data $X$, weight matrix $Q^*$ and label predictor $P^*$.

---

To prove Theorem 1, we use a similar convergence proof method of NMF [4] and CCF [14] by involving an auxiliary function to assist the analysis. We first show the definition of the auxiliary function and its property.

**Definition 1:** $G(x,x')$ is an auxiliary function for $F(x)$ if the following conditions are satisfied:

$$G(x,x') \geq F(x), \quad G(x,x) = F(x) .$$
(49)

**Lemma 1:** If $G$ denotes an auxiliary function, then $F$ is non-increasing under the update:

$$x^{t+1} = \arg\min_x G(x,x') .$$
(50)

**Proof:** $F(x^{t+1}) \leq G(x^{t+1}, x^t) \leq G(x^t, x^t) = F(x^t)$.

Note that the equality $F(x^{t+1}) = F(x^t)$ holds only if $x^t$ is a local minimum of $G(x, x')$. By iterating the above updates, we can easily obtain a sequence of estimates that can converge to a local minimum $x_{\min} = \arg\min_x F(x)$. Next, we define an auxiliary function for our objective function and use Lemma 1 to show that the minimum of the objective function is exactly our update rule, and therefore the Theorem 1 can be proved.

We firstly prove the convergence of the updating rule in Eq.(22). For any entry $w_{ij}$ in $W$, let $F_{w_{ij}}$ be the part of objective function relevant to $w_{ij}$, i.e., Eq.(17). Since the update is essentially element-wise, it is sufficient to show each $F_{w_{ij}}$ is non-increasing under the updating rules. To prove it, we can define the auxiliary function $G$ for $F_{w_{ij}}$.

**Lemma 2:** The following function is an auxiliary function for $F_{w_{ij}}$, which is only relevant to $w_{ij}$:

$$G(w, w_{ij}^t) = F_{w_{ij}}(w_{ij}^t) + F'_{w_{ij}}(w_{ij}^t)(w - w_{ij}^t) + \frac{(KWZ^T A^T AZ)_{ij}}{w_{ij}^t}(w - w_{ij}^t)^2 .$$
(51)

**Proof:** The Taylor series expansion of $F_{w_{ij}}$ is described as

$$F_{w_{ij}}(w) = F_{w_{ij}}(w_{ij}^t) + F'_{w_{ij}}(w_{ij}^t)(w - w_{ij}^t) + 1/2 F''_{w_{ij}}(w - w_{ij}^t)^2 . \quad (52)$$

Note that $\partial^2 \mathcal{O} / \partial W^2 = 2KZ^T A^T AZ$, $F''_{w_{ij}} = (K)_{ii}(Z^T A^T AZ)_{jj}$, and

$$\begin{aligned}(KWZ^TA^TAZ)_{ij} &= \sum_k (KW)_{ik}(Z^TA^TAZ)_{kj}\\ &\geq (KW)_{ij}(Z^TA^TAZ)_{jj} \geq \sum_k (K)_{ik} w^t_{kj}(Z^TA^TAZ)_{jj} \quad ,(53)\\ &\geq w^t_{ij}(K)_{ii}(Z^TA^TAZ)_{jj} \geq w^t_{ij}\frac{1}{2}F''_{w_{ij}}.\end{aligned}$$

where $\mathcal{O}$ denotes the objective function of our RS²ACF. Thus, we can easily conclude that $G(w,w^t_{ij}) \geq F_{w_{ij}}(w)$.

Note that the auxiliary function for the objective function with regard to variable $q_{ij}$ is defined as follows:

**Lemma 3:** The following function $G(q,q^t_{ij})$, where

$$G(q,q^t_{ij}) = F_{q_{ij}}(q^t_{ij}) + F'_{q_{ij}}(q^t_{ij})(q-q^t_{ij}) + \frac{(Y_{new}^T Y_{new}Q)_{ij}}{q^t_{ij}}(q-q^t_{ij})^2$$

is an auxiliary function for $F_{(a_U)_{ij}}$, which is the part of $\mathcal{O}$ that is only relevant to the variable $q_{ij}$.

**Proof:** The proof is essentially similar to that of Lemma 2. By comparing $G(q,q^t_{ij})$ with the Taylor series expansion of $F_{q_{ij}}$, we only need to prove that $(Y_{new}^T Y_{new}Q)_{ij}/q^t_{ij} \geq 1/2 F''_{q_{ij}}$. Because we have $(Y_{new}^T Y_{new}Q)_{ij} = \sum_k (Y_{new}^T Y_{new})_{ik} q_{kj} \geq q^t_{ij}(Y_{new}^T Y_{new})_{ii}$, $\partial^2 \mathcal{O}/\partial Q^2 = 2Y_{new}^T Y_{new}$ and $F''_{q_{ij}} = (Y_{new}^T Y_{new})_{ii}$, we can conclude $(Y_{new}^T Y_{new}Q)_{ij} \geq (q^t_{ij}F''_{q_{ij}})/2$, which can lead to $G(q,q^t_{ij}) \geq F_{q_{ij}}(q)$.

It should be noticed that we can similarly prove that $G(z_L,(z_L)^t_{ij}) \geq F_{(z_L)_{ij}}(z_L)$, $G(z_U,(z_U)^t_{ij}) \geq F_{(z_U)_{ij}}(z_U)$, and $G(a_U,(a_U)^t_{ij}) \geq F_{(a_U)_{ij}}(a_U)$, where $G(w,w^t_{ij})$, $G(z_L,(z_L)^t_{ij})$, $G(z_U,(z_U)^t_{ij})$, $G(a_U,(a_U)^t_{ij})$ and $G(q,q^t_{ij})$ are respectively the auxiliary functions for $F_{w_{ij}}$, $F_{(z_L)_{ij}}$, $F_{(z_U)_{ij}}$, $F_{(a_U)_{ij}}$ and $F_{q_{ij}}$. According to Lemma 1, by computing

$$\begin{aligned}w^{t+1} &= \arg\min_w G(w,w^t_{ij})\\ (z_L)^{t+1} &= \arg\min_{z_L} G(z_L,(z_L)^t_{ij})\\ (z_U)^{t+1} &= \arg\min_{z_U} G(z_U,(z_U)^t_{ij}) \quad ,\\ (a_U)^{t+1} &= \arg\min_{a_U} G(a_U,(a_U)^t_{ij})\\ q^{t+1} &= \arg\min_q G(q,q^t_{ij})\end{aligned} \quad (54)$$

one can similarly obtain the following equations:

$$w^{t+1}_{ij} = w^t_{ij} - w^t_{ij}\frac{F'_{w_{ij}}(w^t_{ij})}{(KWZ^TA^TAZ)_{ij}} = w^t_{ij}\frac{(KAZ)_{ij}}{(KWZ^TA^TAZ)_{ij}}, \quad (55)$$

$$(z_L)^{t+1}_{ij} = (z_L)^t_{ij} - (z_L)^t_{ij}\frac{F'_{(z_L)_{ij}}((z_L)^t_{ij})}{(M_L + \alpha O_L)_{ij}} = (z_L)^t_{ij}\frac{(A_L X_L^T X_L W_L)_{ij}}{(M_L + \alpha O_L)_{ij}}, \quad (56)$$

$$(z_U)^{t+1}_{ij} = (z_U)^t_{ij} - (z_U)^t_{ij}\frac{F'_{(z_U)_{ij}}((z_U)^t_{ij})}{(M_U + \alpha O_U)_{ij}} = (z_U)^t_{ij}\frac{(A_U X_U^T X_U W_U)_{ij}}{(M_U + \alpha O_U)_{ij}}, \quad (57)$$

$$\begin{aligned}(a_U)^{t+1}_{ij} &= (a_U)^t_{ij} - (a_U)^t_{ij}\frac{F'_{(a_U)_{ij}}((a_U)^t_{ij})}{(G_U + \alpha G_L + \beta A_U)_{ij}}\\ &= (a_U)^t_{ij}\frac{(\beta X_U^T P + X_U^T X_U W_U Z_U^T)_{ij}}{(G_U + \alpha G_L + \beta A_U)_{ij}},\end{aligned} \quad (58)$$

$$q^{t+1}_{ij} = q^t_{ij} - q^t_{ij}\frac{F'_{q_{ij}}(q^t_{ij})}{(Y_{new}^T Y_{new}Q)_{ij}} = q^t_{ij}\frac{(Y_{new}^T Y_{new})_{ij}}{(Y_{new}^T Y_{new}Q)_{ij}}, \quad (59)$$

which are exactly the same updates as in Eqs.(22)(33-35)(44) respectively. Thus, the objective function of RS²ACF in Eq.(16) is non-increasing under the updates, which will be verified by quantitative convergence analysis.

### 3.4 Computational Complexity Analysis

We briefly discuss the computational time complexity of our proposed RS²ACF. We use the big $O$ notation to show the complexity of one algorithm as [39]. According to the updating rules of our RS²ACF, we only need to perform the extra updating of $A_U$, $P$, $Q$ and $E$ over the CCF method. Since the big $O$ of each updating operation for each variable is not more than $O(N^3)$ in the optimization procedure of our RS²ACF if $N$ is larger than dimension $D$, the overall time complexity of our RS²ACF is $O(N^3)$.

## 4 RELATIONSHIP ANALYSIS

In this section, we also discuss the important issues that are closely related to our proposed RS²ACF.

### 4.1 Connections with CF [5] and LCCF [10]

Both CF and LCCF aim at finding two matrices $W$, and $V$, where the product of $X$, $W$ and $V$ is the approximation to the data $X$. Recalling the objective function of RS²ACF in Eq.(16), if we substituting $V^T = Z^T A^T$ back, we can have the following simplified framework for our RS²ACF:

$$\begin{aligned}\min_{W,V,Q,A_U,P} J(W,V,Q,A_U,P) &= \|X - E - XWV^T\|_F^2\\ &+ \alpha\left(\|X - XQ\|_F^2 + \|V^T - V^TQ\|_F^2\right) + \beta g(A_U,P)\end{aligned} \quad (60)$$

s.t. $W,V,Q \geq 0$, $Q_{ii} = 0$

It is clear that if $\beta = \gamma = 0$ and the weight matrix $Q$ is fixed in the optimization, the above problem can be reduced to

$$\min_{W,V} J(W,V) = \|X - XWV^T\|_F^2 + \alpha\left(\|X - XQ\|_F^2 + tr(V^T HV)\right), \quad (61)$$

s.t. $W,V \geq 0$

when the sparse error is not included, i.e., $E=0$, where $H = (I-Q)(I-Q)^T$. Note that $\|X - XQ\|_F^2$ is a constant when $Q$ is fixed and the above problem can be equivalent to the objective function of LCCF if $H$ is equivalent to the graph Laplacian $L = \hat{D} - S$ in LCCF, where $\hat{D}$ is a diagonal matrix whose entries are column (or row) sums of the weight matrix $S$. Note that $H = L$ when $S = Q = ee^T/N$, where $e$ is a column vector of all ones. But fixing the weight matrix $Q$ by pre-calculating it before minimizing the reconstruction error will make LCCF lose the adaptive locality preserving ability, and also the pre-calculated weights cannot be ensured to be optimal for subsequent matrix factorization. It is also noticed that if we further constrain $\alpha = 0$, i.e., discarding the local manifold structure preservation power, the above problem just identifies the objective function of CF. Thus, both LCCF and CF are reduced formulations and special cases of our RS²ACF. That is, RS²ACF can potentially outperform both LCCF and CF for data representation and high-dimension data analysis.

### 4.2 Connection with CCF [14]

We discuss the relations between CCF and RS²ACF. One common property is that they all take class information as the additional constraints. Recalling the constraint matrix

*A* of CCF in Eq.(7), the constraint matrix for the unlabeled data is set as an identity matrix, i.e., it can only use class information of labeled data and cannot predict the labels of unlabeled data. In contrast, RS²ACF can clearly obtain a label indicator $A_u$ for unlabeled data to capture the discriminating hidden effects. It is clear that when $\alpha=\beta=0$ in our RS²ACF, we can have the following reduced problem:

$$\min_{W,V,E} J(W,V,E) = \left\| X - E - XWV^T \right\|_F^2 + \gamma \left\| E^T \right\|_{2,1}, \ s.t. \ W, V \geq 0 \ . \ (62)$$

By comparing the problem of CCF with Eq.(62), one can easily find that Eq.(62) is a robust variant of CCF if the same $A_u$ is used. When the error term $E=0$, the problems of CCF and Eq.(62) are completely equivalent. Thus, CCF is clearly a special example of our RS²ACF. But note that setting $\alpha=\beta=0$ means that the reduced problem cannot keep the manifold structures and the discriminant information hidden in unlabeled data cannot be mined to enhance the representation. Setting $E=0$ means that the gross sparse errors in data cannot be modeled any more.

## 5 EXPERIMENTAL RESULTS AND ANALYSIS

We conduct extensive experiments to evaluate RS²ACF for representing, clustering and classifying high-dimensional data. In our study, we mainly compare the performance of our RS²ACF with eight closely related matrix factorization and representation methods, including four unsupervised methods (i.e., NMF, CF, PNMF and LCCF), and four semi-supervised methods (i.e., CNMF, GDNMF, SemiGNMF and CCF). Note that SemiGNMF also adds class information of labeled data into the graph structures by modifying the graph weight matrix [14]. In this study, four public face databases (i.e., JAFFE [18], AR [19], MIT CBCL [20] and UMIST databases [21]), two public object databases (i.e., COIL100 [22] and ETH80 databases [23]), and two public handwritten databases (i.e., CASIA-HWDB1.1 [24][25] and USPS [26]), are used to test the universal applicability of RS²ACF. Detailed information of used databases is shown in Table 1, where we show the total number of samples, dimension and the number of classes. For each face or object image databases, we follow the common evaluation procedures [49-50] to resize all the images into 32×32 pixels for each method for efficiency, and then further convert each image to a 1024-dimensional sample vector with each entry of the vector containing the grey values of image pixels. Finally, we can obtain a data matrix with the vectorized representations of all the images as its columns. Note that the vectorized preprocessing for the handwritten image databases is similar in spirit. We perform all the simulations on a PC with Intel Core i5-4590 CPU @ 3.30 GHz 3.30GHz 8G.

**Table 1:** List of used datasets and dataset information.

| Data Type | Dataset Name | # sample | # dim | # class |
|---|---|---|---|---|
| Face datasets | UMIST [21] | 1012 | 1024 | 20 |
| | JAFFE [18] | 213 | 1024 | 10 |
| | AR [19] | 2600 | 1024 | 100 |
| | MIT CBCL [20] | 3240 | 1024 | 10 |
| Object datasets | COIL100 [22] | 7200 | 1024 | 100 |
| | ETH80 [23] | 3280 | 1024 | 80 |
| Handwritten datasets | CASIA-HWDB1.1-D [25] | 2381 | 196 | 10 |
| | USPS [26] | 9298 | 256 | 10 |

### 5.1 Visualization of the Graph Adjacency Matrix

For representation learning, preserving the local manifold structures of the new representations by weight construction is important for enhancing the representation power. Thus, we would like to compare the constructed adaptive weight matrix $Q$ in our RS²ACF with other two traditional methods to define the graph adjacency matrix, i.e., Gaussian function and LLE-style reconstruction weights [12]. In this study, the MIT face database is applied and we randomly choose 20 images from each class (totally 200 images) to form the matrix $X$ for clear observation, with 10 labeled samples in each class for our RS²ACF method. The kernel width in Gaussian function is defined by using the method of [28] and the number of nearest neighbors is set to 7 [29]. To evaluate the robustness of weight learning to noise, we also prepare a setting under the noisy case.

We visualize the constructed three weight matrices over the original data and noisy data in Figs.2-3, respectively. To corrupt data, we add random Gaussian noise by using $X = X + Variance \times randn(D,N)$ into original data $X$, where the variance is set to 20. Note that we also evaluate the three graph adjacency matrices numerically by comparing the reconstruction error $\varepsilon = \left\| X - X\tilde{Q} \right\|_F^2 / \left\| X \right\|_F^2$, where $\tilde{Q}$ is the graph weight matrix obtained by each weighting method. It is clear that the smaller the reconstruction error $\varepsilon$ is, the better the data reconstruction performance will be, and vice versa. We can find that: 1) The learnt graph adjacency matrix by each weighting approach can have approximate block-diagonal structures. Compared with our weights, there are more wrong inter-class connections in Gaussian weights and LLE-style reconstruction weights, which may result in inaccurate similarity measures and subsequent poor data representations. Specifically, more wrong inter-class connections are produced in the Gaussian weights and LLE-style weights over the noisy case. The main reason for the failure may be because they have to select the number of nearest neighbors or kernel width in Gaussian function, and more importantly they usually fix the number of neighbors for each sample, which is not reasonable, because such operation fails to consider the actual distributions of different real data; 2) In contrast, our RS²ACF computes the adaptive weights jointly without needing to specify the number of nearest neighbors or kernel width used in Gaussian function, so RS²ACF can obtain a weight matrix with less wrong inter-class connections and good intra-class connectivity at the same time, which can be attributed to the adaptive formulation of learning weights, since the learnt weights can be adaptive to different datasets, and we do not need to choose the number of nearest neighbors beforehand and also do not fix it for each data as existing methods; 3) From the quantitative evaluation of reconstruction, we can conclude that our adaptive weight matrix can obtain the smaller reconstruction error than other two weighting methods, i.e., using our adaptive weight matrix to reconstruct the data $X$ is more accurate. As a result, the encoded local manifold structures by the adaptive weight matrix of our RS²ACF would be potentially more accurate and powerful for enhancing the representation ability. The reconstruction error by Gaussian weights is larger than that by LLE-style reconstruction weights in both original and noisy cases. In addition, we see that the reconstruction error of each method over the noisy data is higher than that on original data, which implies that the noise in data can indeed decrease the representation ability of encoded weights. Therefore, learning a robust weight matrix is crucial for data representation.

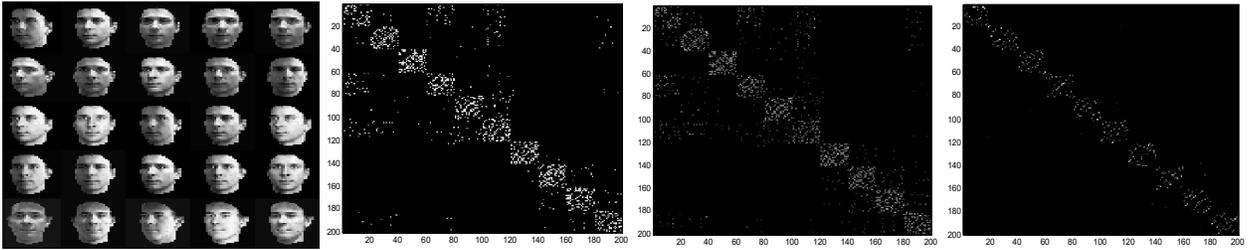

**Quantitative reconstruction errors:** $\|X - X\widetilde{Q}\|_F^2 / \|X\|_F^2 = 20.9677$  $\|X - X\widetilde{Q}\|_F^2 / \|X\|_F^2 = 1.2048$  $\|X - X\widetilde{Q}\|_F^2 / \|X\|_F^2 = 0.1180$

**Fig.2:** Original images *(first)*, and the visualizations of Gaussian weights *(second)*, LLE-style reconstruction weights *(third)* and the adaptive weights in our RS²ACF *(fourth)* on the original MIT face database.

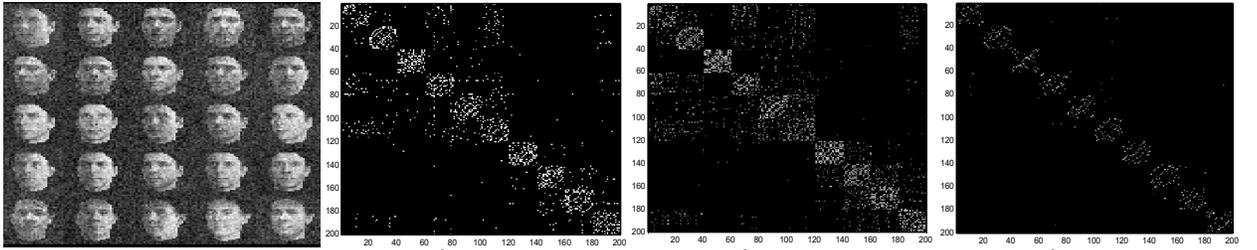

**Quantitative reconstruction errors:** $\|X - X\widetilde{Q}\|_F^2 / \|X\|_F^2 = 23.1403$  $\|X - X\widetilde{Q}\|_F^2 / \|X\|_F^2 = 3.2812$  $\|X - X\widetilde{Q}\|_F^2 / \|X\|_F^2 = 0.2208$

**Fig.3:** Original noisy images *(first)*, and the visualizations of Gaussian weights *(second)*, LLE-style reconstruction weights *(third)* and the adaptive weights in our RS²ACF *(fourth)* on the noisy MIT face database with random corruptions.

## 5.2 Convergence Analysis

We have proved that the objective function value of our RS²ACF in Eq.(16) is non-increasing under the updating rules in Subsection 3.3, so we would like to present some quantitative convergence analysis results to verify it. Two face databases (i.e., MIT CBCL and AR), one object database (i.e., ETH80), and one handwritten digit database (i.e., CASIA-HWDB1.1-D) are evaluated. For each database, we respectively choose 20%, 40%, 60% and 80% labeled data from each class to train our RS²ACF, since we would like to investigate how the percentage of labeled data affect the convergence of our RS²ACF. The convergence results over different percentages of labeled samples are illustrated in Fig.4, where the *x*-axis is the number of iterations and the *y*-axis is the objective function value. We find that: 1) for different percentages of labeled data, RS²ACF can produce the similar convergence trends. That is, the convergence of our RS²ACF is robust to the percentage of labeled samples, because the percentage of labeled data has small effect on the convergence; 2) the objective function values of RS²ACF are non-increasing and also decrease smoothly. Specifically, our RS²ACF with the updating rules converges rapidly and the number of iterations is usually less than 15 in most cases.

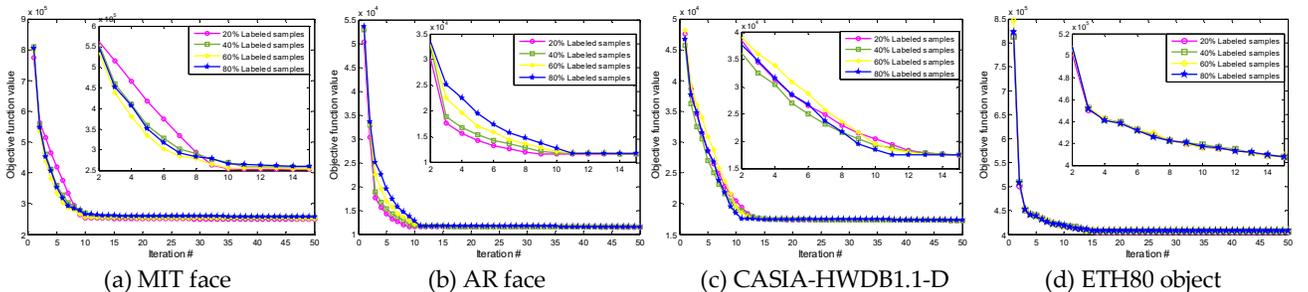

(a) MIT face  (b) AR face  (c) CASIA-HWDB1.1-D  (d) ETH80 object

**Fig.4:** Convergence curves of our RS²ACF algorithm based on four face, object and handwritten digit databases.

## 5.3 Clustering Evaluations

**1) Evaluation Metric:** In this study, we use two widely-used quantitative clustering evaluation methods, i.e., *Accuracy* (AC) and *Normalized Mutual Information* (NMI) [30-31], to evaluate the clustering results. AC is the percentage of the cluster labels to the true labels provided by the original data corpus. The NMI value is computed over the set of clusters obtained from the ground truth and the set of clusters obtained by performing clustering over the new representations by each algorithm. Note that the values of both AC and NMI range from 0 to 1, and the higher the value is, the better the clustering result will be.

**2) Clustering Evaluation Results:** We compare RS²ACF with other methods for data clustering on MIT CBCL and AR face image databases. More specifically, we conduct K-means clustering [32] on the new representation of each algorithm. Following the common evaluation procedures [8-10][14][29], for each fixed cluster number K, we choose K categories from the datasets randomly and mix the data of these K categories as the data matrix *X* for matrix factorization and clustering [29]. Resembling [14], the rank of factorization is set to K+1. In the experiments, we ran-domly choose 30% samples from each class to construct the labeled set for the semi-supervised learning as [14]. For each method, we average the clustering results, i.e., AC and NMI, over 20 times K-means clustering. Note that the clustering results based on varied K values are shown in Tables 2-3. We can draw the following conclusions. 1) The performance of each method goes down as the number of tested categories increases, since clustering data of less categories is relatively easier; 2) The semi-supervised CNMF, SemiGNMF, CCF and RS²ACF methods can deliver better results than the unsupervised NMF, CF, PNMF and LCCF in most cases. More specifically, RS²ACF delivers comparable and even higher accuracies and NMI values than the semi-supervised factorization algorithms, i.e., CNMF, SemiGNMF, GDNMF and CCF.

**Table 2:** Numerical clustering evaluation result of each algorithm on the MIT CBCL face database.

| K | Clustering Accuracy (%) | | | | | | | | | |
|---|---|---|---|---|---|---|---|---|---|---|
| | K-means | NMF | PNMF | CF | LCCF | CNMF | CCF | SemiGNMF | GDNMF | **RS²ACF** |
| 2 | 53.5±1.5 | 54.5±2.0 | 57.8±3.3 | 56.6±1.7 | 72.5±4.5 | 73.5±5.4 | 75.4±4.4 | 81.6±6.4 | 83.5±5.7 | **88.6±5.2** |
| 3 | 46.5±4.2 | 49.3±1.1 | 51.2±2.3 | 50.7±2.6 | 61.8±3.9 | 59.5±5.4 | 57.9±3.2 | 68.9±5.5 | 72.2±4.3 | **84.0±5.8** |
| 4 | 43.9±6.1 | 44.7±0.9 | 48.2±1.3 | 46.5±2.2 | 54.8±3.6 | 55.2±1.1 | 53.4±0.7 | 58.6±5.4 | 62.2±4.2 | **69.5±5.9** |
| 5 | 44.2±1.0 | 46.3±0.5 | 48.8±2.0 | 48.2±0.8 | 53.3±4.2 | 53.4±1.7 | 51.8±4.2 | 57.4±5.0 | 61.0±4.3 | **66.2±5.6** |
| 6 | 38.9±2.0 | 41.2±0.5 | 47.9±1.2 | 39.9±2.5 | 48.2±3.9 | 49.3±1.3 | 47.6±1.4 | 54.2±4.8 | 56.2±4.9 | **62.0±6.3** |
| 7 | 42.3±3.4 | 43.2±1.6 | 43.3±3.1 | 45.6±1.9 | 47.4±3.6 | 51.5±2.2 | 48.8±2.1 | 52.6±3.8 | 54.4±4.5 | **61.2±5.2** |
| 8 | 37.8±2.4 | 38.1±2.5 | 40.2±2.5 | 40.5±2.2 | 47.6±4.2 | 49.5±4.1 | 46.1±2.4 | 53.5±4.2 | 52.9±4.5 | **59.7±4.9** |
| 9 | 37.0±3.5 | 38.6±1.7 | 41.0±2.1 | 38.9±1.3 | 45.5±3.7 | 45.2±3.2 | 40.8±2.9 | 51.0±3.0 | 51.0±4.1 | **57.6±4.7** |
| 10 | 44.5±2.9 | 44.9±4.2 | 46.8±2.3 | 45.2±2.8 | 46.3±3.5 | 51.6±3.1 | 48.2±2.6 | 51.2±3.3 | 50.6±3.7 | **56.9±5.0** |
| K | Normalized Mutual Information (%) | | | | | | | | | |
| | K-means | NMF | PNMF | CF | LCCF | CNMF | CCF | SemiGNMF | GDNMF | **RS²ACF** |
| 2 | 68.5±1.3 | 70.3±2.1 | 73.2±1.4 | 72.5±4.1 | 79.3±4.2 | 78.5±3.9 | 77.2±4.5 | 86.1±6.8 | 88.2±5.5 | **92.5±6.1** |
| 3 | 64.8±3.2 | 66.5±2.6 | 68.7±2.9 | 65.7±2.5 | 73.2±4.1 | 73.6±4.9 | 71.5±3.5 | 83.3±5.9 | 84.9±5.2 | **89.6±5.8** |
| 4 | 60.5±8.5 | 62.2±3.3 | 66.8±2.5 | 65.5±1.8 | 68.0±4.2 | 70.5±3.6 | 68.5±2.5 | 80.5±5.2 | 82.2±5.2 | **88.2±5.2** |
| 5 | 52.5±6.7 | 56.5±1.5 | 59.5±1.1 | 54.6±0.9 | 64.3±1.5 | 68.0±2.5 | 66.3±0.6 | 78.2±4.1 | 79.9±5.4 | **83.5±5.5** |
| 6 | 44.9±4.5 | 46.0±0.8 | 50.3±3.0 | 49.6±1.7 | 62.5±2.8 | 66.6±1.0 | 64.7±3.0 | 72.0±3.2 | 76.0±5.0 | **79.6±4.2** |
| 7 | 58.2±1.7 | 59.6±1.4 | 61.5±2.6 | 60.2±2.1 | 64.5±1.9 | 67.5±1.6 | 65.4±2.2 | 71.1±3.4 | 72.2±5.1 | **78.5±4.0** |
| 8 | 57.1±2.7 | 59.2±1.9 | 62.2±1.5 | 62.3±2.8 | 66.1±2.6 | 66.8±3.1 | 68.3±2.4 | 72.3±3.6 | 70.5±4.5 | **77.2±4.9** |
| 9 | 56.9±3.0 | 58.0±3.2 | 60.6±2.9 | 59.6±2.6 | 64.0±3.0 | 65.3±2.2 | 64.3±2.5 | 70.5±3.8 | 69.8±4.6 | **72.6±4.2** |
| 10 | 54.8±2.4 | 56.1±3.2 | 59.5±2.0 | 58.5±2.2 | 59.1±2.1 | 60.3±2.5 | 60.7±1.8 | 68.4±3.6 | 67.5±4.2 | **68.3±4.1** |

**Table 3:** Numerical clustering evaluation results of each algorithm on the AR face database.

| K | Clustering Accuracy (%) | | | | | | | | | |
|---|---|---|---|---|---|---|---|---|---|---|
| | K-means | NMF | PNMF | CF | LCCF | CNMF | CCF | SemiGNMF | GDNMF | **RS²ACF** |
| 2 | 50.2±0.6 | 57.5±13.8 | 58.0±0.1 | 53.3±3.6 | 62.8±2.5 | 59.8±7.9 | 59.5±3.4 | 72.2±4.0 | 72.3±3.9 | **82.5±8.0** |
| 3 | 36.4±7.8 | 45.6±12.2 | 46.9±0.5 | 39.7±3.1 | 50.5±2.2 | 52.1±7.6 | 49.2±6.9 | **68.8±3.9** | 65.8±3.5 | 66.0±7.4 |
| 4 | 31.8±10.4 | 40.2±7.5 | 44.7±4.7 | 36.3±2.9 | 43.2±1.8 | 45.5±5.3 | 46.3±1.5 | 61.2±3.6 | 60.2±3.8 | **63.8±6.9** |
| 5 | 30.1±6.5 | 35.2±6.6 | 41.2±0.3 | 29.3±1.5 | 41.8±2.0 | 42.1±4.9 | 44.0±3.5 | 55.5±3.6 | 57.5±3.1 | **61.6±6.6** |
| 6 | 26.9±4.7 | 26.5±3.2 | 35.2±4.3 | 25.1±1.8 | 37.3±1.9 | 38.1±3.0 | 37.3±2.3 | 52.6±3.2 | 52.5±3.8 | **54.5±6.2** |
| 7 | 24.3±4.7 | 24.9±2.7 | 27.4±1.8 | 24.7±1.4 | 35.6±1.3 | 31.7±2.5 | 34.7±3.4 | 50.1±3.5 | 48.6±3.2 | **52.5±6.5** |
| 8 | 22.3±3.9 | 23.6±2.5 | 28.8±3.5 | 22.8±1.1 | 33.8±1.6 | 32.9±3.3 | 34.4±3.1 | 45.9±2.9 | 45.2±2.9 | **50.8±6.3** |
| 9 | 21.9±4.3 | 25.6±4.7 | 28.2±3.3 | 22.9±1.0 | 32.5±1.5 | 34.1±4.1 | 35.5±3.1 | 44.5±2.9 | 43.0±3.0 | **50.5±5.9** |
| 10 | 20.6±3.8 | 23.7±3.9 | 27.2±2.5 | 22.2±1.2 | 30.9±1.2 | 32.5±3.4 | 34.2±2.5 | 44.8±3.1 | 42.6±2.8 | **49.2±5.8** |
| K | Normalized Mutual Information (%) | | | | | | | | | |
| | K-means | NMF | PNMF | CF | LCCF | CNMF | CCF | SemiGNMF | GDNMF | **RS²ACF** |
| 2 | 20.1±0.0 | 58.3±17.9 | 22.6±0.6 | 25.3±1.3 | 58.9±3.5 | 59.3±7.8 | 46.2±13.1 | 60.6±5.8 | 62.6±4.8 | **69.2±8.2** |
| 3 | 46.0±14.2 | 46.5±21.6 | 41.1±2.0 | 35.8±2.7 | 56.2±3.4 | 51.2±10.9 | 52.9±10.6 | 62.5±5.2 | 64.5±5.2 | **68.9±8.0** |
| 4 | 42.8±15.8 | 39.7±12.1 | 41.5±5.9 | 36.9±1.5 | 50.2±4.0 | 48.7±6.0 | 46.8±3.2 | 59.2±5.8 | 62.0±4.5 | **65.2±7.6** |
| 5 | 31.6±9.2 | 37.5±9.1 | 35.1±4.1 | 38.3±1.1 | 40.9±2.9 | 42.3±3.9 | 43.5±3.4 | 56.8±5.6 | 59.2±3.8 | **64.4±7.4** |
| 6 | 24.2±6.5 | 24.9±5.3 | 26.8±5.7 | 24.5±1.3 | 36.9±3.2 | 34.7±3.1 | 42.7±2.1 | 52.8±5.1 | 58.2±4.0 | **64.3±6.8** |
| 7 | 29.6±6.4 | 28.5±3.8 | 28.3±2.0 | 28.8±1.5 | 38.2±4.2 | 34.8±2.5 | 42.4±3.0 | 53.4±5.4 | 55.4±3.9 | **62.5±7.1** |
| 8 | 25.2±5.2 | 27.5±2.4 | 26.3±4.1 | 27.3±1.4 | 37.3±3.2 | 34.3±3.5 | 45.6±2.6 | 55.3±4.9 | 54.7±4.3 | **61.6±6.5** |
| 9 | 22.8±6.1 | 32.1±6.7 | 26.7±4.5 | 29.2±1.1 | 36.9±2.9 | 38.4±4.6 | 45.8±3.3 | 51.0±5.0 | 53.8±4.2 | **60.8±6.2** |
| 10 | 22.2±4.9 | 23.1±5.1 | 25.1±3.4 | 22.9±1.0 | 39.2±2.5 | 41.5±3.6 | 42.1±5.1 | 50.8±4.8 | 52.5±4.5 | **60.9±6.8** |

### 5.4 Face Recognition

We first use face recognition to evaluate the distinguishing ability of the learnt new representation by each factorization method. Three face image databases, i.e., JAFFE, MIT CBCL and UMIST, are used for evaluations. The performance of our RS²ACF is mainly compared with those of NMF, CF, PNMF, LCCF, CNMF, SemiGNMF, GDNMF and CCF. Note that the classification process is described as follows. First, we perform each method to compute the new representation. Then, classification is performed over the new representation of each method. To evaluate the results, we use two popular classifiers, i.e., one-nearest-neighbor (1NN) and a semi-supervised classifier termed *Sparse Neighborhood Propagation* (SparseNP) [33]. To avoid the randomness induced by bias, the accuracy is averaged based on 15 times random splits of training/testing images w.r.t. each case and each method for fair comparison.

**Classification on the original images.** We firstly evaluate each algorithm for classifying the original face images. For each semi-supervised method, the number of labeled face images per class is set to 30% of the number of samples. For the face classification by SparseNP, the number of labeled training data of each class is also set to 30% of the total number of training samples. Note that the comparison results by the 1NN and SparseNP classifiers are shown in Fig.5 and Fig.6 respectively, where the horizontal axis is the number of training data per class and the vertical axis denotes the accuracy. We have the following finding. (1) The result of each method is enhanced by increasing the number of training data. (2) RS²ACF delivers better recognition results than other factorization methods across all numbers of training data. Since the recognition task is performed based on the learnt new represen-tation of each method, we can conclude that the obtained representation of RS²ACF is potentially more discriminating than those of NMF, CF, PNMF, LCCF, CNMF, Semi-GNMF, GDNMF and CCF for classification, which can be attributed to the unified model of integrating the adaptive manifold preservation with semi-supervised concept factorization, and enriching the supervised prior knowledge by jointly predicting the labels of unlabeled samples; (3) PNMF, LCCF and the semi-supervised methods can outperform CF and NMF for face recognition in most cases due to the use of labeled data or local information.

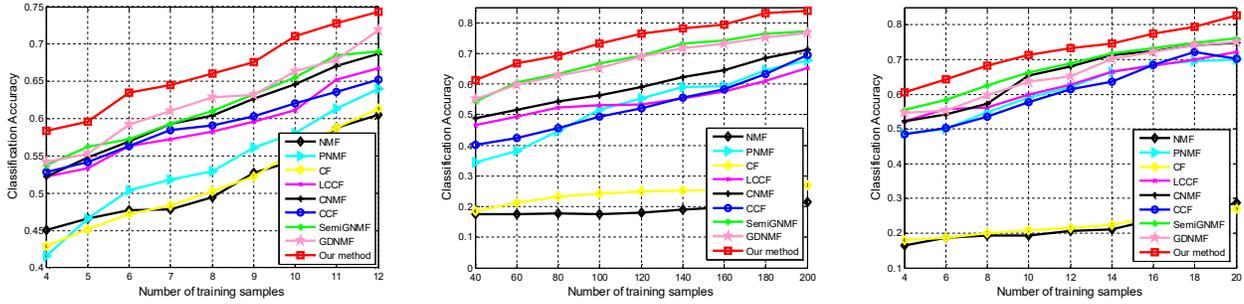

**Fig.5:** Classification Accuracy vs. varied training numbers using 1NN classifier on the original images of JAFFE database *(left)*, MIT database *(middle)*, and UMIST face database *(right)*.

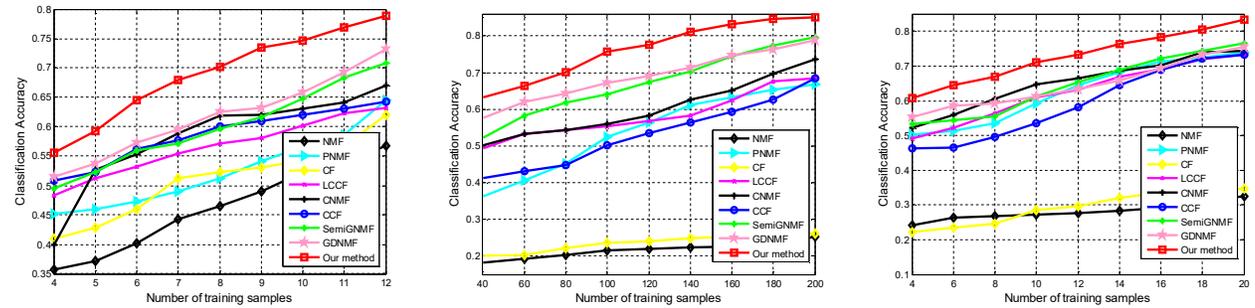

**Fig.6:** Classification Accuracy vs. varied training numbers using SparseNP classifier on the original images of JAFFE database *(left)*, MIT database *(middle)*, and UMIST face database *(right)*.

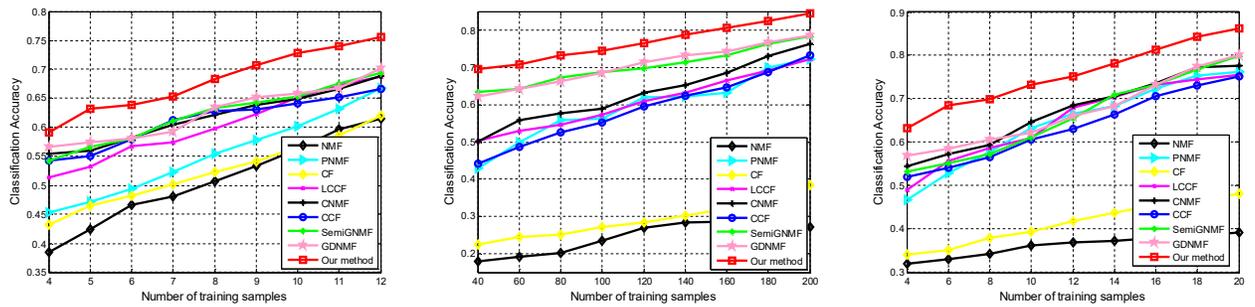

**Fig.7:** Classification Accuracy vs. varied training numbers using 1NN classifier on the random face features of JAFFE database *(left)*, MIT database *(middle)*, and UMIST database *(right)*.

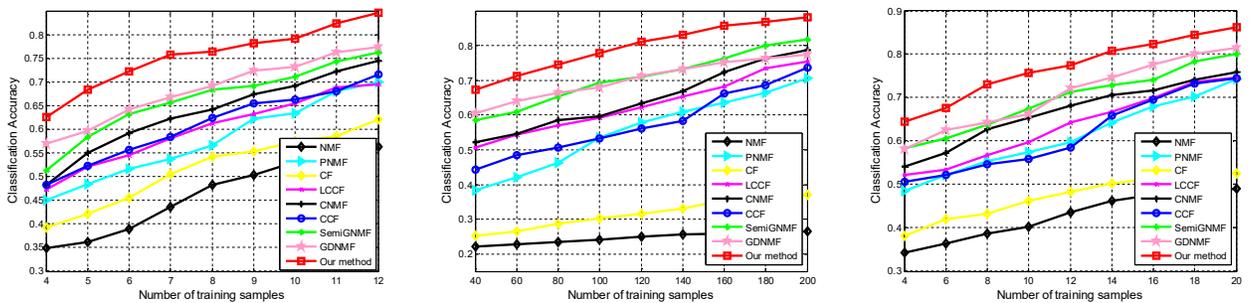

**Fig.8:** Classification Accuracy vs. varied training numbers using SparseNP classifier on the random face features of JAFFE database *(left)*, MIT database *(middle)*, and UMIST database *(right)*.

**Classification on random face features.** We also evaluate each algorithm for classification over the random face features. The random feature descriptor is widely-used for classification [34], [35]. To extract the random features, each image is projected onto a $d$-dimensional feature vector with a randomly generated matrix from a zero-mean normal distribution. Each row of the matrix is L2 normalized. Similar to [34], [35], the dimensionality of a random face image feature is set to $d$=540 in our study. The classification results using the 1NN and SparseNP classifiers on random features are illustrated in Figs.7 and 8, respectively. As can be seen from the figures, similar conclusions regarding the change trends of results and superiority of evaluated methods can be drawn from the results.

### 5.5 Object Recognition

We then conduct object recognition on two popular databases, i.e., ETH80 and COIL100. The result of our RS²ACF is also compared with those of NMF, PNMF, CF, LCCF, CNMF, CCF and SemiGNMF. ETH80 contains images of 8 big categories, and each big category contains 10 subcategories. In this study on ETH80, each big category is treated as a single class, so an eight-class problem is evaluated. For recognition on ETH80, the number of training data in each class varies from {50, 100,…, 300}, and the training number per class tunes from {10, 15,…, 50} for COIL100. For each database, the number of labeled samples is set to 30% of the total training number for the semi-supervised learning. We also provide the results obtained by the 1NN and SparseNP classifiers. The results on the two databases over the varied training numbers are shown in Tables 4 and 5, respectively, where we mainly show the averaged accuracy±standard deviation (%) and the best record (%) for each model, and the highest accuracies in each study are shown in bold. We can find that: (1) Based on the learnt representations of each method, our RS²ACF tends to outperform other recent algorithms. The overall performance of CCF, LCCF, PNMF, CNMF and SemiGNMF are comparable to each other on the ETH80 database due to their different properties. For the COIL100 database, LCCF, CNMF, CCF and SemiGNMF methods can debate PNMF, CF and NMF for recognition. CF and NMF are the worst methods in each setting. (2) Under the same simulation settings, the results by SparseNP are usually higher than those obtained by 1NN, which can be attributed to the fact that SparseNP can use both labeled and unlabeled data for the semi-supervised classification.

**Table 4:** Classification results on the ETH80 database.

| Result / Method | 1NN classifier | | SparseNP classifier | |
|---|---|---|---|---|
| | Mean±std | Best | Mean±std | Best |
| NMF | 40.75±3.24 | 43.83 | 41.67±3.20 | 45.42 |
| PNMF | 63.34±4.20 | 66.75 | 66.85±2.69 | 69.55 |
| CF | 42.05±2.45 | 44.50 | 43.20±3.85 | 46.75 |
| LCCF | 68.22±4.25 | 72.10 | 70.95±4.58 | 74.28 |
| CNMF | 69.85±4.62 | 72.47 | 72.53±3.21 | 76.68 |
| CCF | 65.66±4.30 | 68.98 | 69.06±3.15 | 73.39 |
| SemiGNMF | 69.62±5.20 | 73.21 | 72.55±4.96 | 78.02 |
| GDNMF | 68.95±4.95 | 74.06 | 73.05±5.03 | 79.62 |
| **RS²ACF** | **73.52±5.68** | **77.95** | **77.82±5.54** | **84.10** |

**Table 5:** Classification results on the COIL100 database.

| Result / Method | 1NN classifier | | SparseNP classifier | |
|---|---|---|---|---|
| | Mean±std | Best | Mean±std | Best |
| NMF | 43.02±2.93 | 46.51 | 47.41±4.00 | 52.09 |
| PNMF | 62.58±8.01 | 71.23 | 66.57±8.02 | 75.91 |
| CF | 50.26±4.26 | 54.25 | 52.87±4.20 | 56.13 |
| LCCF | 72.80±2.95 | 74.05 | 74.11±3.53 | 78.32 |
| CNMF | 64.82±8.18 | 73.18 | 65.49±8.65 | 74.56 |
| CCF | 62.62±6.27 | 70.46 | 64.35±6.67 | 73.23 |
| SemiGNMF | 73.16±7.55 | 84.45 | 76.56±6.72 | 85.50 |
| GDNMF | 74.52±5.36 | 83.80 | 79.30±5.88 | 86.38 |
| **RS²ACF** | **77.25±8.20** | **87.52** | **82.25±7.63** | **88.95** |

### 5.6 Handwritten Digit Recognition

We evaluate RS²ACF and other methods for handwritten digit recognition. CASIA-HWDB1.1 and USPS databases are involved. For CASIA-HWDB1.1 [24], the subset called HWDB1.1-D [25], including 2381 handwritten digits ('0'-'9'), from CASIA-HWDB1.1 is used for the evaluation. For USPS database, we choose the first 3000 handwritten digits (i.e., 300 per class) for the simulation. For each dataset, the number of training digits varies from {40, 80,…, 200}. The averaged accuracy rates and best record (%) over different training numbers) by the 1NN and SparseNP classifiers are reported in Tables 6 and 7, from which we can find that RS²ACF still delivers better results than its competitors. CNMF, CCF, SemiGNMF and LCCF obtain better results than NMF, PNMF, and CF in most cases. Thus, RS²ACF can represent the handwritten digits appropriately, i.e., the learnt representations of digits are more discriminative than other methods, which can once again be attributed to the seamless integration of the adaptive local preservation with semi-supervised concept factorization, and the joint learning of class indicator for unlabeled data.

**Table 6:** Classification results on the USPS database.

| Result / Method | 1NN classifier | | SparseNP classifier | |
|---|---|---|---|---|
| | Mean±std | Best | Mean±std | Best |
| NMF | 61.00±2.94 | 64.10 | 63.50±2.25 | 66.30 |
| PNMF | 49.14±0.75 | 55.80 | 53.70±3.21 | 59.51 |
| CF | 52.46±3.03 | 54.10 | 56.05±2.70 | 58.54 |
| LCCF | 72.52±6.32 | 80.05 | 75.55±6.28 | 82.20 |
| CNMF | 72.13±5.66 | 77.80 | 76.58±4.85 | 84.28 |
| CCF | 70.14±5.49 | 79.52 | 73.68±5.20 | 83.75 |
| SemiGNMF | 73.62±6.24 | 81.25 | 76.20±5.97 | 84.49 |
| GDNMF | 74.65±5.44 | 82.24 | 78.28±5.82 | 83.59 |
| **RS²ACF** | **77.59±6.32** | **84.16** | **82.22±5.80** | **87.05** |

**Table 7:** Classification results on the HWDB1.1-D database.

| Result / Method | 1NN classifier | | SparseNP classifier | |
|---|---|---|---|---|
| | Mean±std | Best | Mean±std | Best |
| NMF | 19.55±2.13 | 22.46 | 23.35±2.50 | 26.41 |
| PNMF | 48.21±3.54 | 53.25 | 50.92±2.45 | 56.35 |
| CF | 18.82±2.13 | 22.23 | 21.53±2.31 | 25.57 |
| LCCF | 68.23±3.81 | 70.25 | 70.47±3.35 | 74.80 |
| CNMF | 61.85±4.02 | 64.48 | 64.18±4.70 | 69.35 |
| CCF | 56.49±3.26 | 58.88 | 59.68±4.21 | 64.42 |
| SemiGNMF | 72.85±5.15 | 74.92 | 75.35±5.08 | 79.67 |
| GDNMF | 73.68±5.50 | 77.59 | 74.45±5.66 | 80.73 |
| **RS²ACF** | **77.62±4.59** | **79.84** | **78.39±4.20** | **84.22** |

## 5.7 Parameters Sensitivity Analysis

**(1) Hyperparameter analysis of our RS²ACF.** We mainly explore the effects of the model parameters $\alpha$, $\beta$ and $\gamma$ on the result of RS²ACF. Due to page limitation, the quantitative data classification result is reported as an example, and one face database (i.e., UMIST) is chosen. Since our RS²ACF has three parameters, we first fix $\gamma=10^4$ and use the widely-used grid search strategy [37-38] to tune $\alpha$ and $\beta$ from the candidate set $\{10^{-8}, 10^{-6}, 10^{-4},\ldots, 10^8\}$. Then we fix $\alpha=10^4$ and $\beta=10^{-4}$ to tune $\gamma$ from the candidate set. For classification, the number of training data per class is set to 50% of the total number of samples, and the number of labeled training data of each class is also set to 50% of the number of training samples. The 1NN classifier is used for evaluation in this study. The classification accuracy is averaged over 15 random splits of training/testing samples. The parameter selection results are illustrated in Fig.9. Note that similar observations can be obtained from other databases, so we simply set $\alpha=10^4$, $\beta=10^{-4}$ and $\gamma=10^4$ for the simulations of RS²ACF in this paper.

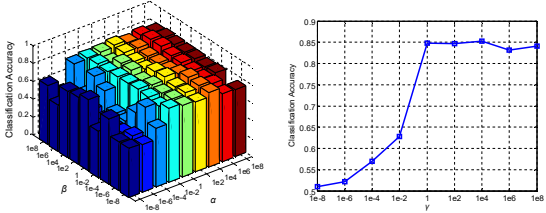

**Fig.9:** Classification results of our RS²ACF over various parameter settings on UMIST face database.

**(2) Hyperparameter analysis of other competitors.** We also present the analysis results of the individual parameters of other competitors on the evaluated databases, i.e., JAFFE, UMIST, MIT, COIL100, ETH80 and USPS databases. Specifically, the evaluated LCCF and SemiGNMF have one parameter in their problems and GDNMF has two parameters. Let $\lambda_{LCCF}$ and $\lambda_{SemiGNMF}$ be the parameters of LCCF and SemiGNMF respectively, and let $\alpha_{GDNMF}$ and $\beta_{GDNMF}$ denote the parameters of GDNMF. We then explore the effects of these parameters on the classification results of the methods. The 1NN classifier is also applied in this study. First, to evaluate LCCF and SemiGNMF, we vary $\lambda$ form the candidate set, and test the classification accuracies under different training samples. Second, to evaluate GDNMF, we also use the grid search strategy [37-38] to tune $\alpha_{GDNMF}$ and $\beta_{GDNMF}$ from candidate set. For JAFFE, UMIST, MIT, COIL100, ETH80, USPS and HWDB1.1-D, 10/20/50/40/50/50/50 samples from each class are chosen as the training set respectively, and the labeled ratio is also set to 30%. The results of LCCF and SemiGNMF are shown in Figs.10-11. The results of GDNMF are shown in Fig.12 as examples, where the number of training samples is set to 50 for MIT, USPS, ETH80 databases, and set to 10/20/40 for JAFFE, UMIST and COIL100 respectively. We can find that LCCF, SemiGNMF and GDNMF obtain the results of similar trends on different databases. Moreover, we report the best choice of hyperparameters in Table 8.

**Table 8.** Settings of parameters for classification by LCCF, SemiGNMF and GDNMF over tested databases.

| Method | JAFFE | UMIST | MIT | COIL100 |
|---|---|---|---|---|
| LCCF ($\lambda$) | $10^6$ | $10^6$ | $10^6$ | $10^6$ |
| SemiGNMF ($\lambda$) | $10^4$ | $10^4$ | $10^4$ | $10^4$ |
| GDNMF ($\alpha$, $\beta$) | ($10^{-6}$, $10^4$) | ($10^{-4}$, $10^{-2}$) | ($10^{-4}$, $10^{-8}$) | ($10^{-4}$, $10^{-6}$) |
| Method | ETH80 | USPS | CASIA-HWDB1.1-D | |
| LCCF ($\lambda$) | $10^8$ | $10^4$ | $10^8$ | |
| SemiGNMF ($\lambda$) | $10^4$ | $10^4$ | $10^4$ | |
| GDNMF ($\alpha$, $\beta$) | ($10^{-6}$, $10^{-4}$) | ($10^{-6}$, $10^2$) | ($10^{-6}$, $10^{-2}$) | |

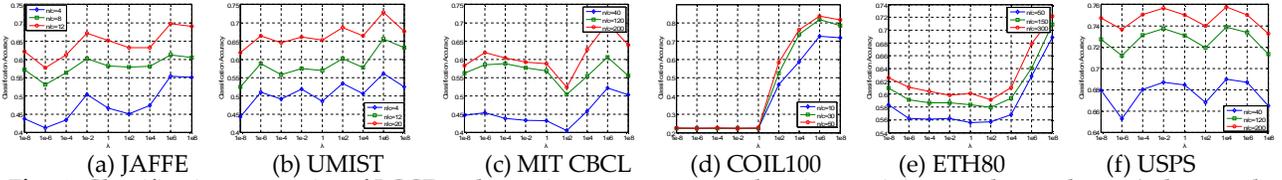

(a) JAFFE    (b) UMIST    (c) MIT CBCL    (d) COIL100    (e) ETH80    (f) USPS

**Fig.10:** Classification accuracies of LCCF under various parameters and various training numbers, where $n/c$ denotes the number of training samples of each class.

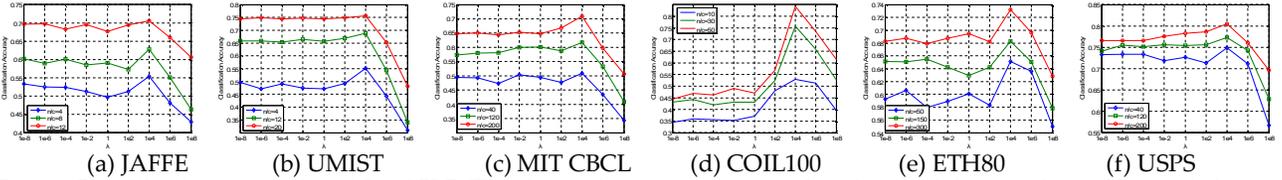

(a) JAFFE    (b) UMIST    (c) MIT CBCL    (d) COIL100    (e) ETH80    (f) USPS

**Fig.11:** Classification accuracies of SemiGNMF under various parameters and various training numbers, where $n/c$ denotes the number of training samples of each class.

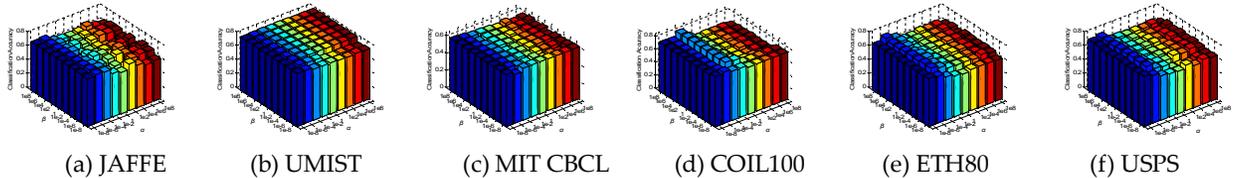

(a) JAFFE    (b) UMIST    (c) MIT CBCL    (d) COIL100    (e) ETH80    (f) USPS

**Fig.12:** Classification accuracies of GDNMF under various parameters.

## 5.8 Further Evaluation on Classification

**(1) Investigation on the rank $r$ of factorization.** In this study, we investigate the effects of rank $r$ on classification. MIT CBCL face database, COIL100 object database and CASIA-HWDB1.1-D handwritten database are evaluated.

For MIT CBCL, we fix the training number to 200, set the number of labeled data to 60 per class and vary $r$ from {4, 6, 8,…,22}. For COIL100, 40 samples are randomly chosen from each class to from the training set with 12 labeled data per class, and the rank $r$ tunes from {15, 30,…,150}. For HWDB1.1-D, the training number per class is set to 100 with 30% labeled, and $r$ is tuned from {4, 6, 8,…,22}.

The analysis results over different rank $r$ of the factorization are shown in Fig.13. We can find that: 1) our RS²ACF delivers higher results than other related models across all choices of $r$; 2) the best results are usually obtained around the point $r=c+1$, where $c$ is the number of classes, which keeps consistent with the conclusion of [14]. Thus, $r=c+1$ is used for each method in all simulations.

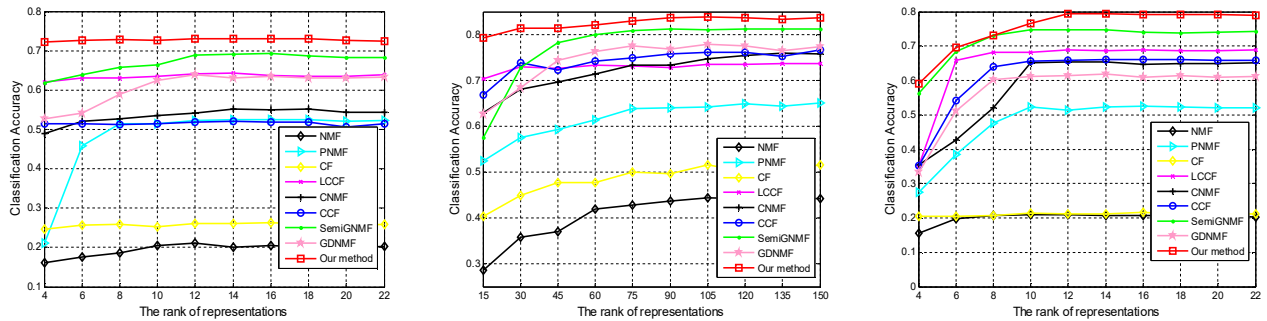

**Fig.13:** Comparison of classification accuracies vs. varied rank $r$ of representations on the MIT CBCL face database (left) and COIL100 object database (middle) and CASIA-HWDB1.1-D database (right), respectively.

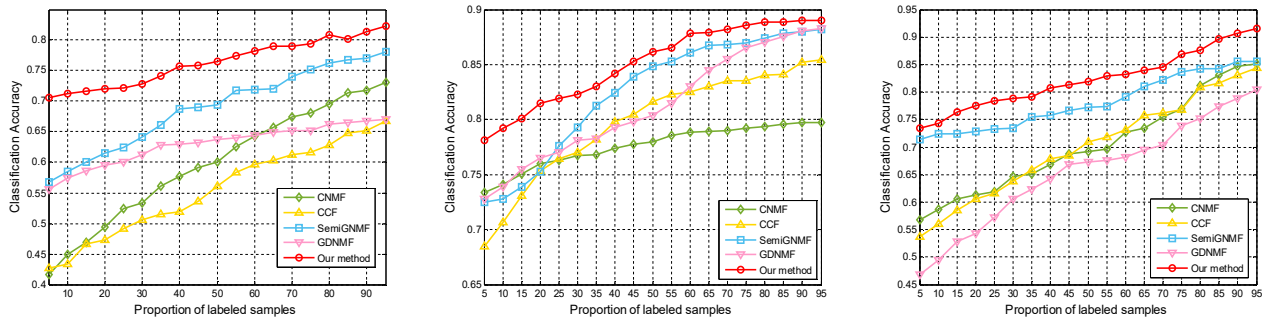

**Fig.14:** Comparison of classification accuracies vs. varied number of labeled samples (per class) on the MIT CBCL face database (left) and COIL100 object database (middle) and CASIA-HWDB1.1-D database (right), respectively.

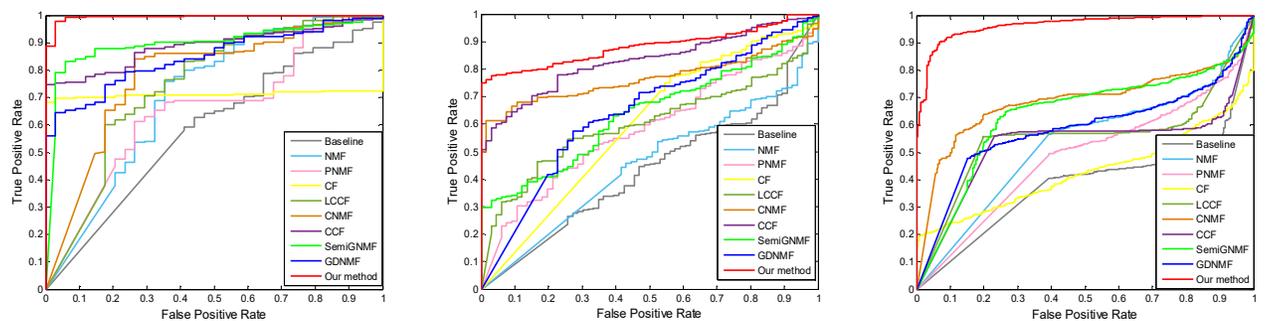

**Fig.15:** ROC curve comparison of each method by the logistic regression classifier on UMIST face database (left), COIL100 object database (middle) and USPS database (right), respectively.

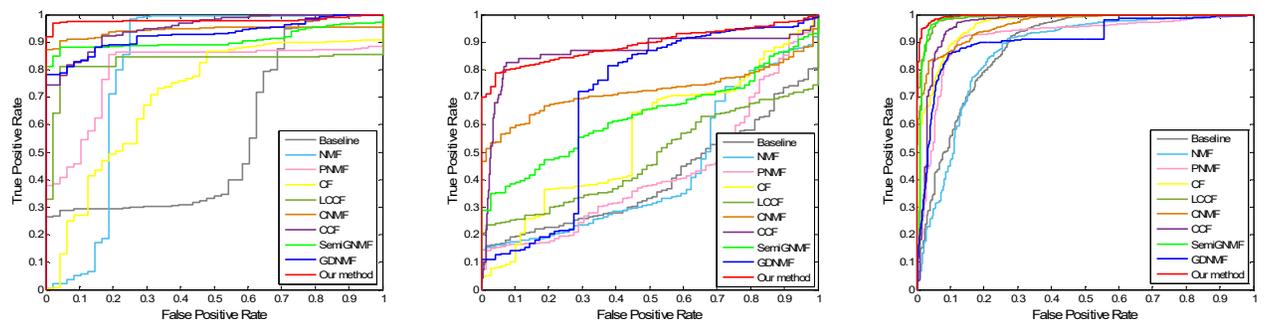

**Fig.16:** ROC curve comparison of each method by the SparseNP classifier on UMIST face database (left), COIL100 object database (middle) and USPS database (right), respectively.

Table 9. AUC and runtime perormance comparison based on three evaluated real-world databases.

| Result / Method | UMIST face database | | | COIL100 object database | | | USPS digit database | | |
|---|---|---|---|---|---|---|---|---|---|
| | AUC1 | AUC2 | Time (s) | AUC1 | AUC2 | Time (s) | AUC1 | AUC2 | Time (s) |
| Baseline | 0.5851 | 0.5548 | - | 0.4317 | 0.4020 | - | 0.3839 | 0.8821 | - |
| NMF | 0.7095 | 0.8089 | 0.56 | 0.4519 | 0.4342 | 1.59 | 0.5329 | 0.8634 | 1.08 |
| PNMF | 0.6534 | 0.8009 | 1.63 | 0.5834 | 0.4209 | 2.25 | 0.4850 | 0.9173 | 8.25 |
| CF | 0.7111 | 0.7082 | 1.55 | 0.5951 | 0.5548 | 40.35 | 0.4278 | 0.9600 | 14.26 |
| LCCF | 0.7467 | 0.8269 | 5.10 | 0.6006 | 0.4762 | 349.64 | 0.5575 | 0.9895 | 39.21 |
| CNMF | 0.7809 | 0.9421 | 10.77 | 0.7663 | 0.7203 | 489.45 | 0.6846 | 0.9629 | 46.68 |
| CCF | 0.8868 | 0.9511 | 14.55 | 0.8285 | 0.8618 | 655.22 | 0.5312 | 0.9610 | 98.96 |
| SemiGNMF | 0.8974 | 0.9082 | 7.82 | 0.6375 | 0.6302 | 38.50 | 0.6356 | 0.9849 | 16.49 |
| GDNMF | 0.8471 | 0.9269 | 6.96 | 0.6442 | 0.6841 | 52.09 | 0.5847 | 0.9085 | 18.50 |
| **Our method** | **0.9938** | **0.9789** | 18.64 | **0.8818** | **0.8912** | 574.35 | **0.9636** | **0.9954** | 84.45 |

**(2) Investigation of the effects of labeled numbers of samples on classification.** In this experiment, five semi-supervised algorithms, i.e., CNMF, GDNMF, SemiGNMF, CCF and our RS$^2$ACF, are evaluated. MIT CBCL, COIL100 and CASIA-HWDB1.1-D databases are also evaluated. In this study, the training number of samples in each class is fixed and the percentage of labeled number of samples is varied. For MIT CBCL, COIL100 and HWDB1.1-D databases, we randomly choose 100, 40, and 100 samples from each class to construct the training sets. The proportion of the labeled samples varies from {5%, 10%,…,95%} for each database. The classification result of each semi-supervised method by the 1NN classifier is shown in Fig.14. We can find: 1) the increasing number of labeled samples can significantly improve the result of each method initially, but the increasing trend becomes slow as the number of labeled data is increased to a higher level; 2) our RS$^2$ACF can deliver the better classification results over different numbers of labeled samples in most cases, and the performance superiority of our RS$^2$ACF over other methods is more obvious when the labeled number is relatively small. Note that this phenomenon is promising for semi-supervised learning, because the number of labeled samples is limited in most real applications.

**(3) Investigation on binary classification.** Binary data classification problem is also discussed. In this study, the binary logistic regression classifier [51] and SparseNP [33] are employed to evaluate the results. The running time performance of each factorization methods is also presented. Three real image databases, i.e., UMIST face database, COIL100 object database and USPS database, are used as examples. For each database, we consider the first class as the positive class data and the other classes are treated as the opposite class data. For UMIST, COIL100 and USPS, we choose 20, 6 and 50 samples from each class to form the training sets respectively, where 6, 3 and 20 samples are selected as labeled. To quantify the binary classification, ROC curves [42-43] and the corresponding area under the ROC curve (AUC) values are shown in Figs.15-16 and Table 9 respectively, where AUC1 denotes the result by the logistic regression classifier and AUC2 is the result by SparseNP. Note that we also report the classification results of the SparseNP and the logistic regression classifier on the original raw data (without going through factorization) as the baseline. To test the runtime, the averaged training time of factorization by each method are reported for fair comparison. We can find that: (1) RS$^2$ACF is superior to other methods by delivering higher AUC values in investigated cases; (2) the results of baseline methods on original raw data are worse than other methods by performing factorization for new representations, which implies that the learnt new representation by factorization can indeed improve the classification; (3) for runtime performance, RS$^2$ACF needs comparable time to CCF, and both needs more time than other methods; (4) logistic regression is on-par with SparseNP in some cases, e.g., AUC1 and AUC2 are comparable on COIL100, while SparseNP is superior to the logistic regression on UMIST and USPS by delivering higher AUC values in most cases.

# 6 CONCLUDING REMARKS

We proposed a new and robust semi-supervised adaptive concept factorization algorithm that aim at improving the discriminating ability of the new representations and the robustness properties to noise and gross sparse errors. To enhance the discriminating power, RS$^2$ACF incorporates class information of labeled data as hard constraints and estimates class information of the unlabeled data by soft constraints. Specifically, RS$^2$ACF jointly obtains a robust label predictor to propagate class information of labeled data to unlabeled data, and also compute an explicit label indicator for unlabeled data. That is, RS$^2$ACF also makes full use of unlabeled data by predicting their labels, and adds them into the constraint matrix as soft constraints to make the representations more discriminative. To encode the locality more accurately, RS$^2$ACF clearly preserves the manifold structures of the labeled and unlabeled data adaptively in the representation space and label space, i.e., the weights are ensured to be optimal for representation and classification of high-dimensional data jointly.

We have conducted extensive clustering and classification simulations to show the effectiveness of RS$^2$ACF. The investigated cases show that enhanced results can be delivered by RS$^2$ACF, compared with other closely related factorization methods. In future, how to choose the optimal rank of factorization needs further investigation. In addition, how to speed up the optimization process of our RS$^2$ACF will also be investigated in future.

## ACKNOWLEDGEMENTS

We want to express our sincere gratitude to anonymous referee and their comments that make our manuscript a higher standard. This work is partially supported by the

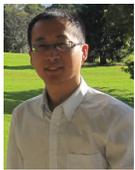

**Zhao Zhang** (SM'17- ) received the Ph.D. degree from the Department of Electronic Engineering (EE), City University of Hong Kong, in 2013. He is now a Full Professor at the School of Computer Science & School of Artificial Intelligence, Hefei University of Technology, Hefei, China. Dr. Zhang was a Visiting Research Engineer at National University of Singapore, worked with Prof. Shuicheng Yan, from Feb to May 2012. He then visited the National Laboratory of Pattern Recognition (NLPR) at Chinese Academy of Sciences, worked with Prof. Cheng-Lin Liu, from Sep to Dec 2012. During Oct 2013 and Oct 2018, he was an Associate Professor at the School of Computer Science and Technology, Soochow University, Suzhou, China. His current research interests include Multimedia Data Mining & Machine Learning, Image Processing & Pattern Recognition. He has authored/co-authored over 80 technical papers published at prestigious journals and conferences. Specifically, he has published 15 regular papers in IEEE/ACM Transactions journals as the first-author/corresponding author. He is serving as an Associate Editor (AE) of Neurocomputing, IEEE Access and IET Image Processing. Besides, he is serving/served as a Senior Program Committee (SPC) member of PAKDD 2019/2018/2017, an Area Chair (AC) of BMVC 2018/2016/2015、ICTAI 2018, a PC member for some popular international conferences (such as CVPR, IJCAI, AAAI, ICDM, SDM, ICPR, PRICAI, PAKDD), and often got invited as a journal reviewer for IEEE/ACM Transactions journals, etc. He is now a Senior Member of the IEEE, and a Member of the ACM.

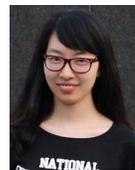

**Yan Zhang** is now working toward the PHD degree at the School of Computer Science and Technology, Soochow University, Suzhou, China, co-supervised by Dr. Zhao Zhang. Her current research interests mainly include data mining, machine learning and pattern recognition. Specifically, she is interested in high-dimensional data analysis and feature extraction. During her PhD study, she has already published papers in the IEEE TKDE and Pattern Recognition.

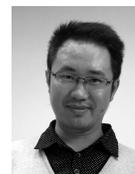

**Guangcan Liu (M'11- )** received the bachelor's degree in mathematics and the Ph.D. degree in computer science and engineering from Shanghai Jiao Tong University, China, in 2004 and 2010, respectively. He was a Post-Doctoral Researcher with the National University of Singapore, Singapore, from 2011 to 2012, the University of Illinois at Urbana-Champaign, Champaign, IL, USA, from 2012 to 2013, Cornell University, Ithaca, NY, USA, from 2013 to 2014, and Rutgers University, Piscataway, NJ, USA, in 2014. Since 2014, he has been a Professor with the School of Information and Control, Nanjing University of Information Science and Technology, Nanjing, China. His research interests are pattern recognition and signal processing. He obtained the National Excellent Youth Fund in 2016 and was designated as the global Highly Cited Researchers in 2017.

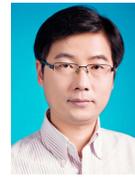

**Jinhui Tang** is a Professor in the School of Computer Science and Engineering, Nanjing University of Science and Technology, China. His current research interests include multimedia search and computer vision. He has authored more than 100 journal and conference papers in these areas. He is a co-recipient of the Best Student Paper Award in MMM 2016, and Best Paper Awards in ACM MM 2007, PCM 2011, and ICIMCS 2011. He serves as an Associate Editor of IEEE Trans. on Knowledge and Data Engineering (TKDE), IEEE Trans. on Neural Networks and Learning Systems (TNNLS) and Information Sciences, as a Technical Committee Member for approximately 30 international conferences, and as a Reviewer for approximately 30 prestigious international journals. He was a co-recipient of the Best Paper Award from ACM Multimedia 2007, PCM 2011, and ICIMCS 2011. He is a senior member of the IEEE and a member of the ACM.

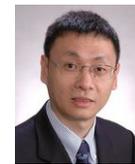

**Shuicheng Yan** (F'16- ) received the Ph.D. degree from the School of Mathematical Sciences, Peking University, in 2004. He is currently the Dean's Chair Associate Professor at National University of Singapore, and also the chief scientist of Qihoo/360 company. Dr. Yan's research areas include machine learning, computer vision and multimedia, and he has authored/ co-authored hundreds of technical papers over a wide range of research topics, with Google Scholar citation over 20,000 times and H-index 66. He is ISI Highly-cited Researcher of 2014-2016. He is an associate editor of IEEE Trans. Knowledge and Data Engineering, IEEE Trans. on Circuits and Systems for Video Technology (IEEE TCSVT) and ACM Trans. Intelligent Systems and Technology (ACM TIST). He received the Best Paper Awards from ACM MM'12 (demo), ACM MM'10, ICME'10 and ICIMCS'09, the winner prizes of classification task in PASCAL VOC 2010-2012, the winner prize of the segmentation task in PASCAL VOC 2012, 2010 TCSVT Best Associate Editor (BAE) Award, 2010 Young Faculty Research Award, 2011 Singapore Young Scientist Award and 2012 NUS Young Researcher Award. He is a Fellow of the IEEE and IAPR.

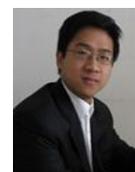

**Meng Wang** is a Professor in the Hefei University of Technology, China. He received the B.E. degree and Ph.D. degree in the Special Class for the Gifted Young and signal and information processing from the University of Science and Technology of China (USTC), Hefei, China, respectively. His current research interests include multimedia content analysis, search, mining, recommendation, and large-scale computing. He has authored 6 book chapters and over 100 journal and conference papers in these areas, including IEEE TMM, TNNLS, TCSVT, TIP, TOMCCAP, ACM MM, WWW, SIGIR, ICDM, etc. He received the paper awards from ACM MM 2009 (Best Paper Award), ACM MM 2010 (Best Paper Award), MMM 2010 (Best Paper Award), ICIMCS 2012 (Best Paper Award), ACM MM 2012 (Best Demo Award), ICDM 2014 (Best Student Paper Award), PCM 2015 (Best Paper Award), SIGIR 2015 (Best Paper Honorable Mention), IEEE TMM 2015 (Best Paper Honorable Mention), and IEEE TMM 2016 (Best Paper Honorable Mention). He is the recipient of ACM SIGMM Rising Star Award 2014. He is/has been an Associate Editor of IEEE Transactions on Knowledge and Data Engineering (TKDE), IEEE Transactions on Neural Networks and Learning Systems (TNNLS) and IEEE Transactions on Circuits and Systems for Video Technology (TCSVT). He is a senior member of the IEEE.